\documentclass{article}

\usepackage{microtype}
\usepackage{graphicx}
\usepackage{subcaption}
\usepackage{booktabs} % for professional tables

\usepackage{hyperref}

\usepackage{xcolor}         % colors
\usepackage{multirow}
\usepackage{url}            % simple URL typesetting

\usepackage{amsmath,amsfonts,bm,mathrsfs}

\def\eqref#1{equation~\ref{#1}}
\def\1{\bm{1}}

\def\vtheta{{\bm{\theta}}}

\def\va{{\bm{a}}}

\def\ve{{\bm{e}}}

\def\vg{{\bm{g}}}

\def\mA{{\bm{A}}}
\def\mB{{\bm{B}}}

\def\mF{{\bm{F}}}
\def\mG{{\bm{G}}}

\def\mU{{\bm{U}}}

\def\mX{{\bm{X}}}

\DeclareMathAlphabet{\mathsfit}{\encodingdefault}{\sfdefault}{m}{sl}
\SetMathAlphabet{\mathsfit}{bold}{\encodingdefault}{\sfdefault}{bx}{n}

\def\gB{{\mathcal{B}}}

\def\gL{{\mathcal{L}}}

\def\gO{{\mathcal{O}}}

\newcommand{\E}{\mathbb{E}}

\newcommand{\R}{\mathbb{R}}
\newcommand{\T}{\top}

\newcommand{\pardiff}[1]{\frac{\partial}{\partial#1}}

\definecolor{matpltblue}{HTML}{1f77b4}
\definecolor{matpltorange}{HTML}{ff7f0e}
\definecolor{matpltgreen}{HTML}{2ca02c}
\definecolor{matpltpurple}{HTML}{9467bd}
\definecolor{matpltyellow}{HTML}{bcbd22}

\newcommand{\Curvature}{\textcolor{matpltgreen}{Curvature}}

\newcommand{\Precondition}{\textcolor{matpltyellow}{Precondition}}
\newcommand{\forward}{\textcolor{matpltblue}{forward}}
\newcommand{\backward}{\textcolor{matpltorange}{backward}}
\newcommand{\curvature}{\textcolor{matpltgreen}{curvature}}
\newcommand{\inverse}{\textcolor{matpltpurple}{inverse}}
\newcommand{\inversion}{\textcolor{matpltpurple}{inversion}}
\newcommand{\Inversion}{\textcolor{matpltpurple}{Inversion}}
\newcommand{\precondition}{\textcolor{matpltyellow}{precondition}}
\newcommand{\preconditioned}{\textcolor{matpltyellow}{preconditioned}}
\newcommand{\preconditioning}{\textcolor{matpltyellow}{preconditioning}}

\newcommand{\pldepth}{D}
\newcommand{\plwidth}{W}
\newcommand{\microbs}{B_{micro}}
\newcommand{\nmicrobatches}{N_{micro}}
\newcommand{\minibs}{B_{mini}}

\newcommand{\memoryparam}{M_\theta}
\newcommand{\memoryact}{M_{act}}
\newcommand{\peakmemoryerr}{M_{err}^{peak}}
\newcommand{\savememoryerr}{M_{err}^{save}}
\newcommand{\memorypipe}{M_{pipe}}

\newcommand{\memorykfac}{M_{kfac}^+}
\newcommand{\timekfac}{T_{kfac}^+}
\newcommand{\memorycurv}{M_{\textcolor{matpltgreen}{curv}}}
\newcommand{\timecurv}{T_{\textcolor{matpltgreen}{curv}}}
\newcommand{\memoryinv}{M_{\textcolor{matpltpurple}{inv}}}
\newcommand{\timeinv}{T_{\textcolor{matpltpurple}{inv}}}

\newcommand{\timeprec}{T_{\textcolor{matpltyellow}{prec}}}

\newcommand{\timepipe}{T_{pipe}}
\newcommand{\timebubble}{T_{bubble}}
\newcommand{\timefwd}{T_{\textcolor{matpltblue}{f}}}
\newcommand{\timebwd}{T_{\textcolor{matpltorange}{b}}}
\newcommand{\countfwd}{C_{\textcolor{matpltblue}{f}}}
\newcommand{\countbwd}{C_{\textcolor{matpltorange}{b}}}

\newcommand{\embdim}{d_{model}}
\newcommand{\ffdim}{d_{ff}}
\newcommand{\nheads}{h}
\newcommand{\seqlen}{S}
\newcommand{\params}{\vtheta}

\newcommand{\act}{\va}

\newcommand{\err}{\ve}

\newcommand{\nparams}{P}
\newcommand{\minibatch}{\gB}
\newcommand{\avg}[1]{\left\langle#1\right\rangle_{i\in\minibatch}}
\newcommand{\loss}{\gL}

\newcommand{\nlayers}{L}
\newcommand{\din}{d^{in}}
\newcommand{\dout}{d^{out}}

\newcommand{\fisher}{\mF}
\newcommand{\empfisher}{\hat{\mF}}

\newcommand{\specialcelll}[2][c]{%
  \begin{tabular}[#1]{@{}l@{}}#2\end{tabular}}

\newcommand{\change}[1]{{#1}}

\usepackage[accepted]{mlsys2023}

\mlsystitlerunning{PipeFisher: Efficient Training of Large Language Models Using Pipelining and Fisher Information Matrices}

\begin{document}

\twocolumn[
\mlsystitle{PipeFisher: Efficient Training of Large Language Models Using \\ Pipelining and Fisher Information Matrices}

% It is OKAY to include author information, even for blind
% submissions: the style file will automatically remove it for you
% unless you've provided the [accepted] option to the mlsys2023
% package.

% List of affiliations: The first argument should be a (short)
% identifier you will use later to specify author affiliations
% Academic affiliations should list Department, University, City, Region, Country
% Industry affiliations should list Company, City, Region, Country

% You can specify symbols, otherwise they are numbered in order.
% Ideally, you should not use this facility. Affiliations will be numbered
% in order of appearance and this is the preferred way.
%\mlsyssetsymbol{equal}{*}

\begin{mlsysauthorlist}
\mlsysauthor{Kazuki Osawa}{eth}
\mlsysauthor{Shigang Li}{bupt}
\mlsysauthor{Torsten Hoefler}{eth}
\end{mlsysauthorlist}

\mlsysaffiliation{eth}{Department of Computer Science, ETH Zurich, Switzerland}
\mlsysaffiliation{bupt}{Beijing University of Posts and Telecommunications, China}

\mlsyscorrespondingauthor{Kazuki Osawa, Shigang Li}{kazuki.osawa@inf.ethz.ch, shigangli.cs@gmail.com}

% You may provide any keywords that you
% find helpful for describing your paper; these are used to populate
% the "keywords" metadata in the PDF but will not be shown in the document
\mlsyskeywords{Machine Learning, MLSys}

\vskip 0.3in

\begin{abstract}
    Pipeline parallelism enables efficient training of Large Language Models (LLMs) on large-scale distributed accelerator clusters. Yet, pipeline \textit{bubbles} during startup and tear-down reduce the utilization of accelerators. 
    Although efficient pipeline schemes with micro-batching and bidirectional pipelines have been proposed to maximize utilization, a significant number of bubbles cannot be filled using synchronous forward and backward passes.
    To address this problem, we suggest that \textit{extra work} be assigned to the bubbles to gain \textit{auxiliary benefits} in LLM training. 
    As an example in this direction, we propose \textit{PipeFisher}, which assigns the work of K-FAC, a second-order optimization method based on the Fisher information matrix, to the bubbles to \textit{accelerate convergence}.
    In Phase 1 pretraining of BERT-Base and -Large models, PipeFisher reduces the (simulated) training time to 50-75\% compared to training with a first-order optimizer by greatly improving the accelerator utilization and benefiting from the improved convergence by K-FAC. 
\end{abstract}
]

% this must go after the closing bracket ] following \twocolumn[ ...

% This command actually creates the footnote in the first column
% listing the affiliations and the copyright notice.
% The command takes one argument, which is text to display at the start of the footnote.
% The \mlsysEqualContribution command is standard text for equal contribution.
% Remove it (just {}) if you do not need this facility.

\printAffiliationsAndNotice{}  % leave blank if no need to mention equal contribution
%\printAffiliationsAndNotice{\mlsysEqualContribution} % otherwise use the standard text.

\section{Introduction}
Transformer-based~\cite{vaswani_attention_2017} large language models (LLMs) \cite{devlin_bert_2019,brown_language_2020} are pushing the limits of the capabilities of deep neural network models in a variety of domains.
Since increasing the size of the model is one key factor to increasing the capacity of the LM, massively parallel accelerators (e.g., GPUs, TPUs) are being utilized to speed up the training process.
Simple \textit{data parallelism}, where each accelerator has a copy of the entire model and performs forward and backward passes for a subset of a mini-batch, i.e., \textit{micro-batch} (or local mini-batch), is not feasible for LLMs that do not fit the memory of a single accelerator.
Therefore, it is common in LLM training to combine data parallelism and \textit{model partitioning}, where the model is distributed to multiple accelerators.
%\htor{fix para grammar}

Typical approaches to model partitioning are (i) \textit{operator parallelism}\footnote{Operator parallelism is often also referred to as tensor parallelism \cite{narayanan_efficient_2021} or model parallelism \cite{chowdhery_palm_2022}}, (ii) \textit{state partitioning} (e.g., ZeRO \cite{rajbhandari_zero_2020} for optimizer state and model parameters), and (iii) \textit{pipeline parallelism} \cite{griewank_algorithm_2000,chen_training_2016}. 
Each scheme requires collective communication (allreduce) of intermediate representations (i.e., activations, error signals), collective communication (broadcast) of a partition of the model parameters, and point-to-point (P2P) communication (send/recv) of intermediate representations, respectively.
In (i) operator parallelism and (ii) state partitioning, increasing parallelism results in larger communication overhead.
On the other hand, in the case of (iii) pipeline parallelism, the communication overhead in LLMs is negligible because the P2P communication is small and can easily be overlapped with forward and backward passes, but pipelining creates \textit{bubbles} of time in which accelerators become idle.
Thus, all approaches have overhead, and the one that achieves the highest throughput (number of tokens processed per unit time) depends on various aspects such as the number of parallel accelerators, model size, and interconnect performance.

\begin{figure*}
    \centering
    \includegraphics[width=.85\textwidth]{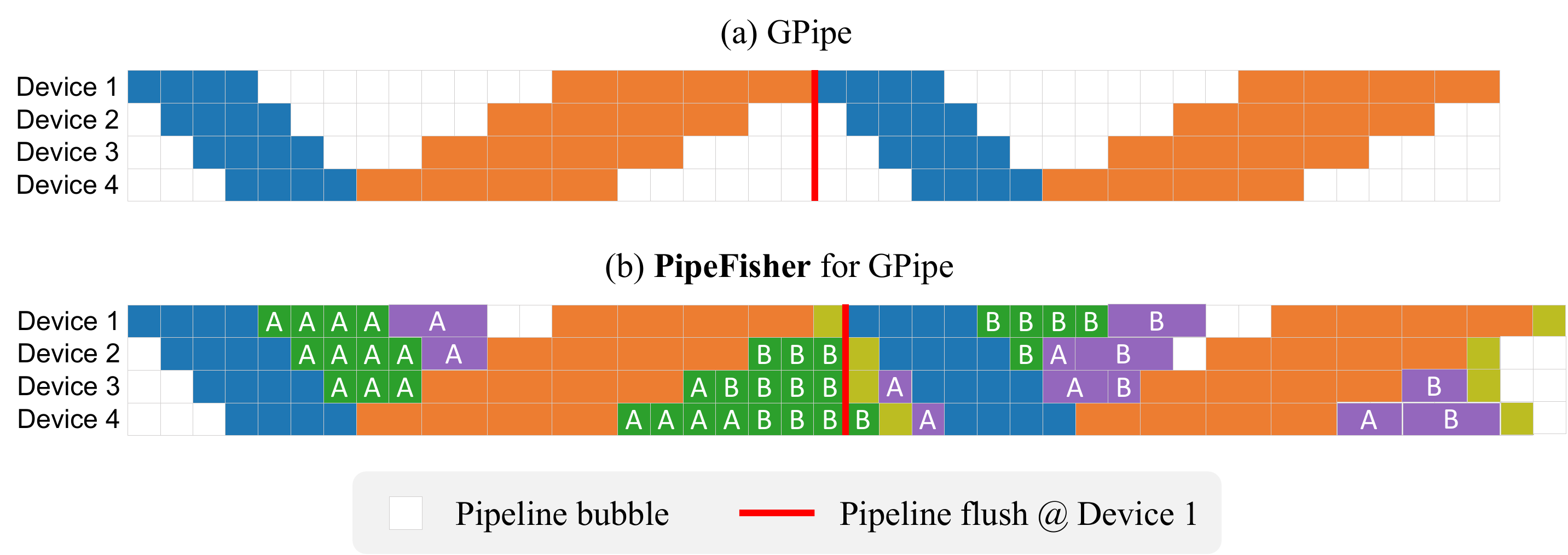}
    \caption{Schematic pipeline schedule (for two steps) of (a) GPipe \cite{huang_gpipe_2019} and (b) \textbf{PipeFisher} for GPipe w/ 4 stages, 4 micro-batches, and 4 devices.
    Each colored box represents a work of {\forward} (for a micro-batch), {\backward} (for a micro-batch), {\curvature} (for $\mA_l$ or $\mB_l$ of a micro-batch), {\inversion} (for $\mA_l$ or $\mB_l$ of (a subset of) assigned layers), or {\precondition}.
    PipeFisher utilizes the pipeline bubbles of multiple pipeline steps (two steps in this schedule) to refresh the {\curvature} and {\inverse} matrices once.
    Thus, {\precondition} is the only computational overhead of PipeFisher over the standard pipeline schemes.
    The first {\precondition} in this schedule is performed with the stale {\inverse} matrices calculated at previous steps.
    }
    \label{fig:pipeline}
\end{figure*}

We note that unlike the other model partitioning approaches, the overhead of pipelining mainly comes from the \textbf{low utilization of the accelerators} rather than communication costs. 
To increase the utilization and throughput, efficient pipeline methods, e.g., GPipe \cite{huang_gpipe_2019}, 1F1B \cite{narayanan_pipedream_2019}, and Chimera \cite{li_chimera_2021}, have been proposed. 
%and asynchronous methods, e.g., PipeDream \cite{narayanan_pipedream_2019} and PipeDream-2BW \cite{narayanan_memory-efficient_2021},
%\htor{what about async pipes - no bubbles (no convergence either but papers claim so ;-))}
It is possible to fill most bubbles with these methods when the number of micro-batches per accelerator is large enough.
However, pipeline parallelism is often combined with data parallelism on massively parallel accelerators to achieve highest throughput \cite{rajbhandari_zero-infinity_2021,narayanan_efficient_2021}.
As a result, not enough micro-batches can be allocated to each accelerator to fill the bubbles efficiently.

In this work, we suggest to assign \textit{extra work} (computation and communication) to the bubbles of pipelines to gain \textit{auxiliary benefits} in LLM training in massively parallel settings. 
Auxiliary benefits in exchange for the extra work include avoidance of the \textit{catastrophic forgetting} in learning through weight- or/and function-space regularizers \cite{kirkpatrick_overcoming_2017,pan_continual_2020}, model compression based on weights and gradient magnitude \cite{evci_rigging_2020}, and improved generalization performance by estimating the loss landscape \cite{foret_sharpness-aware_2021} and avoiding \textit{sharp minima} \cite{hochreiter_flat_1997,keskar_large-batch_2017}.
As an example in this direction whose benefits are relatively easy to observe and has reasonably complex work, we choose \textit{second-order optimization}, which brings us the benefit of \textit{improved convergence} and speeds up LLM training. 
And we propose \textbf{PipeFisher}, a training scheme that \textit{automatically} assigns the work of K-FAC \cite{martens_optimizing_2015}, a second-order optimization method based on the Fisher information matrix, to the bubbles in \textit{any} pipeline schedule. 
\autoref{fig:pipeline} illustrates the pipeline schedule in our PipeFisher method for GPipe \cite{huang_gpipe_2019}.
In Phase1 pretraining of BERT-Base and -Large, PipeFisher \textbf{improves the GPU utilization} in Chimera \cite{li_chimera_2021}, a state-of-the-art pipeline method, \textbf{from 75.9\% to 93.2\%} and \textbf{from 59.8\% to 97.6\%} and reduces the (simulated) training time to 48.7\% and 75.7\% compared to NVLAMB with Chimera, respectively.
%\htor{what are these two options?}

\section{Background and Related Work}

\begin{figure*}[t]
    \centering
    \includegraphics[width=.85\textwidth]{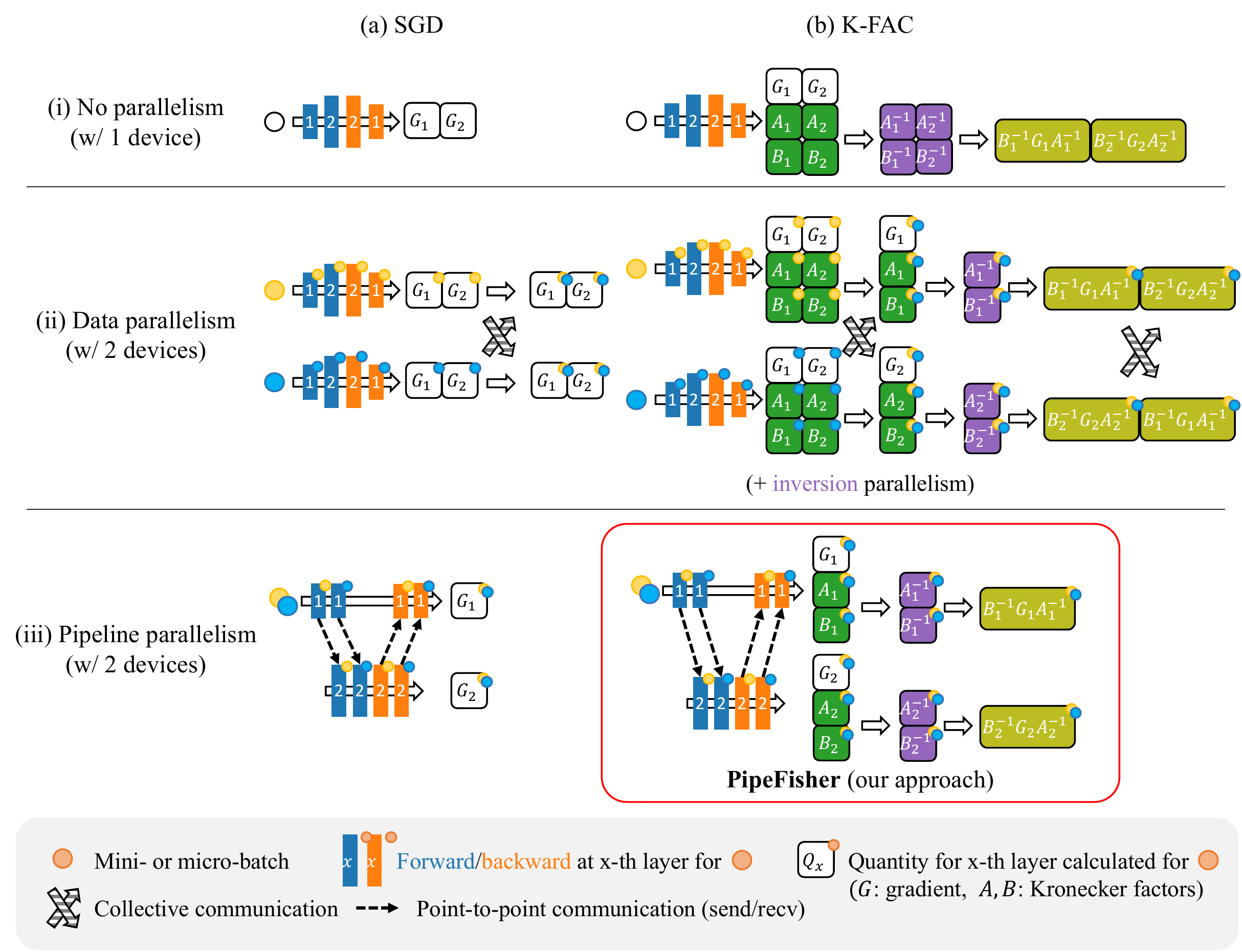}
    \caption{A gradient calculation (and preconditioning) step of (a) SGD and (b) K-FAC using (i) no parallelism, (ii) data parallelism (w/ 2 devices), and (iii) \textbf{pipeline parallelism} (w/ 2 stages, 1 layer/stage, 2 micro-batches, and 2 devices) for a two-layer neural network. 
    In no-parallel and data-parallel K-FAC (i,b and ii,b), {\curvature} (and collective communication after it) and {\inverse} are usually performed once in many steps (e.g., 100 steps \cite{pauloski_kaisa_2021}) to reduce the computational (and communication) overheads. 
    Our \textbf{pipeline-parallel K-FAC} (\textbf{PipeFisher}) (iii,b) performs {\curvature} and {\inverse} in pipeline bubbles and refreshes the matrices once in a few steps (see \autoref{fig:pipeline}, \autoref{fig:gpipe_1f1b_timeline}, and \autoref{fig:chimera_timeline}). 
    For every scheme of K-FAC, {\precondition} is performed every step with fresh or stale {\inverse} matrices ($\mA_l^{-1},\mB_l^{-1}$) for fresh gradients ($\mG_l$).
    }
    \label{fig:parallel}
\end{figure*}

In mini-batch-based training methods, the neural network model receives a mini-batch $\minibatch=\{(x_i,y_i)\}$ of input example and target output, and the mini-batch loss is often evaluated as the average of per-example negative log likelihood of the target output given the input:
\begin{align*}
    \loss_\minibatch(\params)
    :=
    \frac{1}{|\minibatch|}\sum_{i=1}^{|\minibatch|}
    -\log p_\params(y_i|x_i)
    =
    \avg{-\log p_\params(y_i|x_i)}
    \,,
\end{align*}
where $\params\in\R^\nparams$ is a column vector of the model parameters, and $\avg{\cdot}$ represents the average over the mini-batch $\minibatch$.
The conditional probability $p_\params(y_i|x_i)$ is calculated by performing a {\forward} pass on the model followed by a softmax. 
The mini-batch gradient $\vg:=\pardiff{\params}\loss_\minibatch(\params)\in\R^\nparams$ is calculated by a {\backward} pass and is used for the parameter update by gradient-based optimizers (e.g., SGD) (\autoref{fig:parallel}(i,a)).

\subsection{Distributed Parallel Deep Learning}
%There exist several schemes for increasing the throughput (number of examples processed per unit time) of the {\forward} and {\backward} workloads by utilizing distributed accelerators in parallel.

\textbf{Data parallelism}: 
%To increase the mini-batch size $|\minibatch|$ \htor{is that to increase the size, really?}, 
To increase the throughput (number of examples processed per unit time) of the {\forward} and {\backward} work, a mini-batch is sharded across multiple accelerators. 
Each accelerator has a copy of the identical model and performs {\forward} and {\backward} for a different shard of the mini-batch, i.e., \textit{micro-batch} (or local mini-batch.)
To keep model parameters common across the accelerators throughout the training, micro-batch gradients are synchronized through collective communication (i.e., allreduce) at every optimization step (\autoref{fig:parallel}(ii,a)).

\textbf{Pipeline parallelism}:
For a large model that does not fit the memory of an accelerator, the model is divided into multiple partitions or \textit{stages} (sequences of the layers) and each accelerator performs {\forward} and {\backward} on the assigned stage in a pipeline.
In synchronous pipeline methods, at the beginning and end of the pipeline (startup and tear-down), a stage needs to wait for the previous (or the next) stage's {\forward} (or {\backward}) to complete, and there will be \textit{bubbles} of time when some accelerators are idle. 
To better utilize the bubbles, it is a common approach to divide a mini-batch into multiple micro-batches \cite{huang_gpipe_2019,narayanan_pipedream_2019} and overlap the {\forward} (or {\backward}) work on different accelerators (\autoref{fig:parallel}(iii,a)). 

\subsection{Natural Gradient Descent and Fisher Information Matrix}

Natural gradient descent (NGD) \cite{amari_natural_1998} finds the steepest direction with respect to the Kullback-Leibler (KL) divergence between the model's predictive distributions before and after the parameter update: 
\begin{align*}
%    \Delta_{\rm NGD}
%    =
    \mathrm{min}_{\Delta\in\R^\nparams}
    \loss_\minibatch(\params+\Delta)
    \mbox{   s.t.   }
    KL(p_\params||p_{\params+\Delta})=const.
\end{align*}
(cf. gradient descent finds the steepest direction with respect to the Euclidean distance between the parameters.) 
The constraint ensures that the model moves along the manifold of probability distributions at a constant rate \cite{pascanu_revisiting_2014}.
Assuming $\Delta\rightarrow \bf{0}$, $KL(p_\params||p_{\params+\Delta})\approx\frac{1}{2}\Delta^\T\fisher\Delta$ (the second-order Taylor expansion), where $\fisher\in\R^{\nparams\times\nparams}$ is the \textit{Fisher information matrix} (FIM):
\begin{align}
    \label{eq:fim}
    \fisher:=
    \E_{q}\left[\E_{p_\params}
    \left[
    \pardiff{\params}\log p_\params(y|x)
    \pardiff{\params}\log p_\params(y|x)^\T
    \right]\right]
    \,,
\end{align}
where $q$ is the input distribution,
and we get the update direction $\Delta_{\rm NGD}\approx-\fisher^{-1}\vg$.
When the expectation $\E_{q}$ is replaced with the average over the minibatch $\minibatch$, the FIM is equivalent to the generalized Gauss-Newton approximation \cite{pearlmutter_fast_1994} (positive semidefinite approximation) of the Hessian of the mini-batch loss $\loss_\minibatch$, and NGD can be seen as an approximate second-order optimization method \cite{pascanu_revisiting_2014,martens_new_2020}.
In practice of deep learning, the FIM is often estimated by the \textit{empirical Fisher} \cite{martens_new_2020}:
\begin{align*}
    \empfisher:=
    \avg{
    \pardiff{\params}\log p_\params(y_i|x_i)
    \pardiff{\params}\log p_\params(y_i|x_i)^\T
    }
%    \in\R^{\nparams\times\nparams}
    \,,
\end{align*}
where both the expectation $\E_{q}$ and $\E_{p_\params}$ in \autoref{eq:fim} are replaced with the average over the mini-batch $\minibatch$.
This allows the estimate of the FIM to be calculated during the backpropagation for the mini-batch loss $\loss_\minibatch$, leading to a faster training time \cite{osawa_scalable_2022}.

Yet, the NGD is infeasible for deep neural network models with a large number of parameters ($\nparams$ can be from millions to trillions \cite{brown_language_2020}) due to the huge computational and memory cost for constructing and inverting the (estimate of the) FIM (a $\nparams\times\nparams$ matrix).

\subsection{Kronecker-Factored Approximate Curvature (K-FAC)}
To make NGD practical, K-FAC \cite{martens_optimizing_2015} approximates the curvature matrix in NGD (i.e., FIM) with an easy-to-invert matrix.
Here we describe the K-FAC method for $\nlayers$-layer fully-connected neural network (ignoring the biases for simplicity).
The (empirical) FIM is first approximated with a (1) layer-wise block-diagonal matrix (\textit{layer independence}):
$
    \empfisher
    \approx
    \mathrm{block\_diag}
    \left(
        \begin{array}{cccc}
            \empfisher_1 & \empfisher_2 & \cdots & \empfisher_\nlayers
        \end{array}
    \right)
$
 where $\empfisher_l\in\R^{\nparams_l\times\nparams_l}$ ($l=1,\dots,\nlayers$, $\sum_{l=1}^\nlayers\nparams_l=\nparams$) is the FIM for the parameters of the $l$-th layer.
Then (2) the \textit{Kronecker factorization} (\textit{input-output independence}) is applied to each diagonal block:
\begin{align*}
    \empfisher_l
    &=
    \avg{
    \pardiff{\params_l}\log p_\params(y_i|x_i)
    \pardiff{\params_l}\log p_\params(y_i|x_i)^\T
    }
    \\
    &=
    \avg{
    \left(
    \act_{l}^{(i)}
    \otimes
    \err_{l}^{(i)}
    \right)
    \left(
    {\act_{l}^{(i)}}
    \otimes
    {\err_{l}^{(i)}}
    \right)^\T
    }
    \\
    &=
    \avg{
    \act_{l}^{(i)}
    {\act_{l}^{(i)}}^\T
    \otimes
    \err_{l}^{(i)}
    {\err_{l}^{(i)}}^\T
    }
    \\
    &\approx
    \underbrace{
    \avg{
    \act_{l}^{(i)}{\act_{l}^{(i)}}^\top
    }
    }_{=:\mA_l\in\R^{\din_l\times\din_l}}
    \otimes
    \underbrace{
    \avg{
    \err_l^{(i)}{\err_l^{(i)}}^\top
    }}_{=:\mB_l\in\R^{\dout_l\times\dout_l}}
    \in\R^{\nparams_l\times\nparams_l}
    \,.
\end{align*}
$\act_l^{(i)}\in\R^{\din_l\times 1}$ is the input to the $l$-th layer (activations from the previous layer) for the $i$-th example, $\err_l^{(i)}\in\R^{\dout_l\times 1}$ is the gradient of $\loss_\minibatch$ with respect to the outputs (errors) of the $l$-th layer for the $i$-th example, and $\params_l\in\R^{\nparams_l}$ ($\nparams_l=\din_l\cdot\dout_l$) is the set of  parameters of the $l$-th layer ($l=1,\dots,\nlayers$).  
$\otimes$ represents the Kronecker product of two matrices (or vectors): for $\mA\in\R^{d_1\times d_2}$ and $\mB\in\R^{d_3\times d_4}$,
\begin{align*}
\mA\otimes\mB
:=
\left[
    \begin{array}{ccc}
        A_{1,1}\mB & \cdots & A_{1,d_2}\mB \\
        \vdots & \ddots & \vdots \\
        A_{d_1,1}\mB & \cdots & A_{d_1,d_2}\mB
    \end{array}
\right]
\in\R^{d_1d_3\times d_2d_4}
\,.
\end{align*}

\subsubsection{Work in K-FAC}
Besides the {\forward} and {\backward} computations for calculating gradients,
K-FAC requires additional work per optimization step (\autoref{fig:parallel}(i,b)).

\textbf{{\Curvature} work}: 
The Kronecker factors $\mA_l\in\R^{\din_l\times\din_l}$ and $\mB_l\in\R^{\dout_l\times\dout_l}$ ($l=1,\dots,L$) can be calculated by concatenating per-example activations and errors and performing matrix-matrix multiplications: 
\begin{align*}
    \mU_{A,l}&:=\frac{1}{\sqrt{|\minibatch|}}
    \left[
    \begin{array}{ccc}
    \act_{l}^{(1)} & \cdots & \act_{l}^{(|\minibatch|)}
    \end{array}
    \right]
    \in\R^{\din_l\times|\minibatch|}
    \,,
    \\
    \mU_{B,l}&:=\frac{1}{\sqrt{|\minibatch|}}
    \left[
    \begin{array}{ccc}
    \err_l^{(1)} & \cdots & \err_l^{(|\minibatch|)}
    \end{array}
    \right]
    \in\R^{\dout_l\times|\minibatch|}
    \,,
    \\
    \mA_l&=\mU_{A,l}\mU_{A,l}^\T
    \,,
    \mbox{ and }
    \mB_l=\mU_{B,l}\mU_{B,l}^\T
    \,.
\end{align*}
In PyTorch \cite{paszke_pytorch_2019}, this can be implemented by calling \texttt{torch.matmul()} for each of $\mA_l$ and $\mB_l$ for every layer ($2\times\nlayers$ calls in total).

\textbf{{\Inversion} and {\precondition} work}: 
After constructing the Kronecker factors $\mA_l$ and $\mB_l$, we can get the approximate layer-wise natural gradient as follows:
\begin{align*}
    \empfisher_l^{-1}\vg_l
    \approx
    \left(\mA_l\otimes\mB_l\right)^{-1}\vg_l
    &=
    \left(\mA_l^{-1}\otimes\mB_l^{-1}\right)\vg_l
    \\
    &=
    \mathrm{vec}
    \left(
    \mB_l^{-1}
    \mG_l
    \mA_l^{-1}
    \right)
    \in\R^{\nparams_l}
    \,,
\end{align*}
where $\vg_l\in\R^{\nparams_l}$ is the gradient of the loss with respect to the parameters of the $l$-th layer ($l=1,\dots,\nlayers$, $\vg=\left[\begin{array}{ccc}\vg_1^\T & \cdots & \vg_L^\T\end{array}\right]^\T$).
Here we exploit two properties of a Kronecker product:
$\left(\mA\otimes\mB\right)^{-1}=\mA^{-1}\otimes\mB^{-1}$ and
$(\mA\otimes\mB)\mathrm{vec}\left(\mX\right)=\mathrm{vec}\left(\mB\mX\mA\right)$.
$\mathrm{vec}(\cdot)$ is an operator that vectorizes a matrix by stacking its columns, and $\vg_l=\mathrm{vec}(\mG_l)$.
This reduces the computational complexity for the inversion from $\gO(\nparams_l^3)$ to $\gO({\din_l}^3+{\dout_l}^3)$. 
Also, we can avoid the memory consumption for materializing the Kronecker product $\mA_l\otimes\mB_l$ and its {\inverse}.
As each Kronecker factor is a symmetric matrix, we can calculate its {\inverse} by utilizing Cholesky decomposition.
In PyTorch, we call \texttt{torch.linalg.cholesky()} followed by \texttt{torch.linalg.cholesky\_inverse()} for every Kronecker factor ($2\times L$ calls in total).
Finally, we call \texttt{torch.matmul} two times ($2\times\nlayers$ calls in total) to get the {\preconditioned} gradient $\mB_l^{-1}\mG_l\mA_l^{-1}$ ($l=1,\dots,\nlayers$).

In practice, {\curvature} and {\inversion} work are performed only once in many optimization steps, depending on the model size, data size, and available computing resource (e.g., 10 steps for {\curvature} and 100 steps for {\inversion} for pretraining BERT-Large in \cite{pauloski_kaisa_2021}) to reduce the computational overheads of K-FAC. 
%\htor{I find this still very confusing that this works? Is it applied in each iteration or only in the iterations where it's computed (fresh)?}
In this case, {\precondition} of the gradients of the current optimization step will be performed using the stale {\inverse} matrices that are calculated at the previous steps.
%\htor{this is exactly the weird part - oh well! Reason why it's weird is that 2nd order has ``higher resolution'' in my intuition, and now we take the low res SGD info and add outdated high-res 2nd order info - weird}

%\htor{here, I raise the question if we use async 2nd order info, why can we not just use an async pipeline to fill the bubbles and be done with it much simpler (same utilization, maybe even higher) - the downside may be needing more data but we can run multiple epochs - do we have a good argument here?}

\subsubsection{Distributed Parallel K-FAC Schemes}

\textbf{CPU offloading}: 
\citet{ba_distributed_2017} proposed a distributed parallel K-FAC scheme that has multiple \textit{gradient workers} for {\forward} and {\backward} work (data parallelism), a \textit{stats worker} for {\curvature} and {\inversion} work, and a \textit{parameter server} for {\preconditioning} and updating the parameters. 
The stats worker asynchronously calculates $\mA_l,\mB_l,\mA_l^{-1},$ and $\mB_l^{-1}$ ($l=1,\dots,\nlayers$) for a mini-batch on a CPU while the gradient workers process multiple mini-batches on accelerators (GPUs).
Once the {\inverse} matrices are ready, they are sent to the parameter server and are used for {\preconditioning}. 
\citet{anil_scalable_2021} also adopt CPU offloading of {\curvature} and {\inversion} work for a distributed version of Shampoo optimizer \cite{gupta_shampoo_2018} which requires to construct and invert the Kronecker-factored AdaGrad \cite{duchi_adaptive_2011} matrix (i.e., second moment matrix of mini-batch gradients) of the same shapes as the Kronecker-factored FIM in K-FAC.
Because constructing and inverting matrices on CPUs can be much slower than a {\forward} and a {\backward} work on accelerators, the {\inverse} matrices used for {\preconditioning} can be stale for many steps (e.g., 100-1000 steps).
In this scheme, the frequency of refreshing the {\inverse} matrices is bounded by the CPU performance compared to the accelerators.
%\htor{again, we argue staleness wrt performance but not what happens to the data science}

\textbf{Data and {\inversion} parallelism}: \citet{osawa_large-scale_2019} proposed a hybrid scheme of data parallelism and \textit{{\inversion} parallelism} where each accelerator performs {\forward}, {\backward}, and {\curvature} work for a different micro-batch (data parallelism) and performs {\inversion} and {\preconditioning} work for different layers ({\inversion} parallelism) (illustrated in \autoref{fig:parallel}(ii,b)). 
%\htor{I think we're really over/mis-using the team ``workload'' - it's really ``calculation'' or something like this imo} 
This approach efficiently reduces the per-step computational and memory costs of K-FAC, which mainly come from the {\inversion} work, and scales to as many distributed accelerators as the number of layers in the model.
Yet, the {\curvature} work for all the layers need to be performed by each accelerator.
Moreover, this scheme introduces an additional work, i.e., the communication of dense matrices (Kronecker factors of each layer) among distributed accelerators due to the data parallelism, and this will be the main bottleneck in massively parallel settings \cite{ueno_rich_2020}.
To mitigate these computational and communication costs per step, it is a common strategy to apply a manually selected frequency (e.g., once in 100 steps) for refreshing the {\inverse} matrices \cite{pauloski_convolutional_2020,pauloski_kaisa_2021}.  

\section{PipeFisher}
\label{sec:pipefisher}
We propose \textit{PipeFisher}\footnote{\url{https://github.com/kazukiosawa/pipe-fisher}}, a training scheme that assigns the K-FAC work, i.e., {\curvature}, {\inverse}, and {\precondition}, to bubbles in pipelines (\autoref{fig:pipeline} and \autoref{fig:parallel}(iii,b)).
PipeFisher has several advantage over the CPU-offloading and data- and {\inversion}-parallel K-FAC (\autoref{fig:parallel}(ii,b)): 
(i) each accelerator only needs to store the parameters, gradients, and {\curvature} matrices for the layers in the assigned pipeline stage, resulting in smaller memory consumption;
(ii) the {\inverse} work are split among multiple accelerators without collective communication;
(iii) because PipeFisher leverages bubbles to perform {\curvature} and {\inverse} work, these computations do not affect training throughput; 
and (iv) since these computations are performed on accelerators, which is much faster than on CPUs, the matrices can be refreshed much more frequently (e.g., once in 2-3 steps).
This is expected to allow for more stable convergence and more aggressive learning rates since it is observed that the value of curvature matrix fluctuates greatly, especially in the early phase of the training \cite{osawa_large-scale_2019}.
%\htor{this is interesting - would be a nice ablation study just from a data science perspective (on the refresh rate) - not sure we can do it now but can we cite something?}

\subsection{Automatic Work Assignments}

\begin{figure*}
    \centering
    \includegraphics[width=0.8\textwidth]{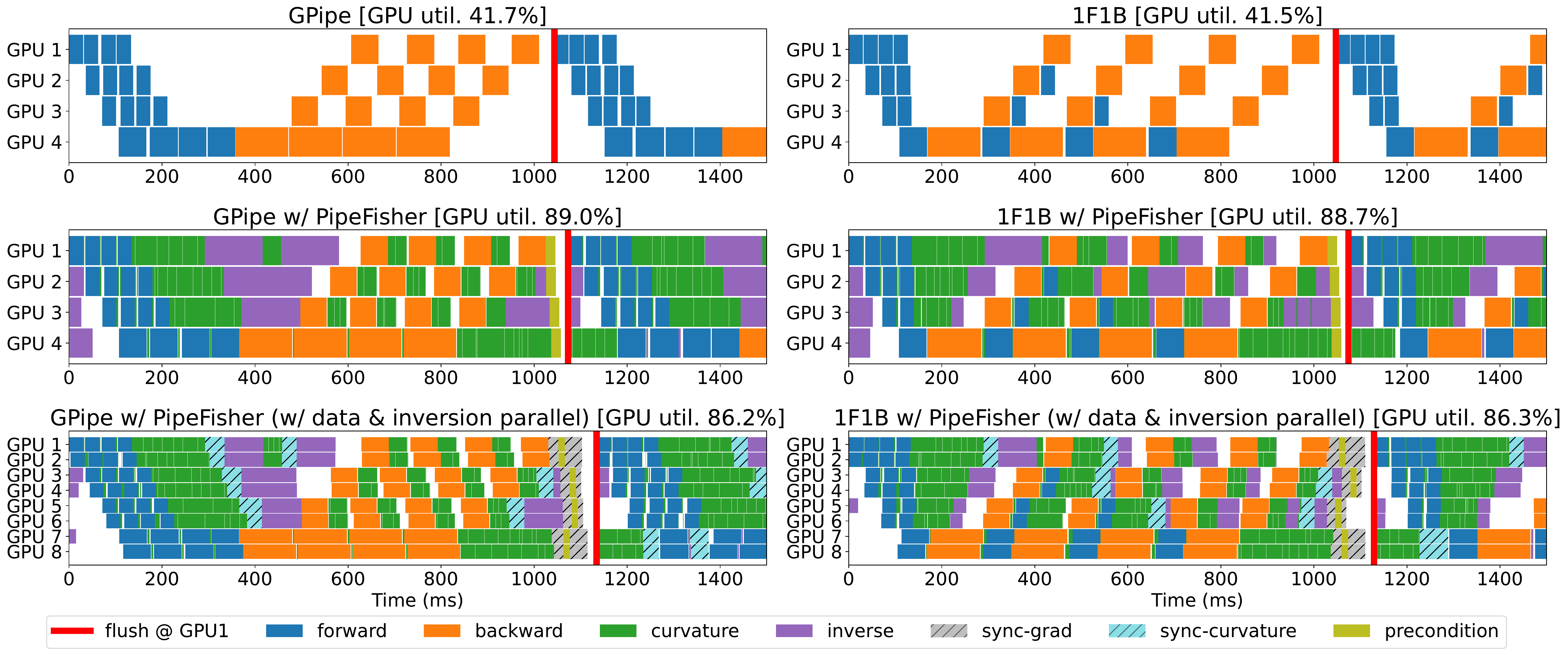}
    \caption{Profiled CUDA kernel execution times on NVIDIA P100 GPUs for 5th and (part of) 6th pipeline steps in GPipe (left) and 1F1B (right) w/ Adam (top) and w/ PipeFisher w/o (middle) and w/ (bottom) data and {\inversion} parallelism in pretraining BERT-Base ($\nlayers=12$) w/ 4 stages (3 layers/stage), 4 or 8 GPUs, 4 micro-batches of size 32 per GPU per step, and sequence length 128.}
    \label{fig:gpipe_1f1b_timeline}
\end{figure*}

%\htor{sure, but how important is this frequency?}
Our goal is to refresh the {\curvature} and {\inverse} matrices as frequently as possible  by utilizing the bubbles (idle accelerators) in \textit{any} pipeline schedule (e.g., GPipe, 1F1B, Chimera) as much as possible.
To this end, PipeFisher \textit{automatically} assigns K-FAC work to bubbles using several rules:
\begin{enumerate}
    \item A {\curvature} work for $\mA_l$ or $\mB_l$ (for a micro-batch and for a layer) is assigned to a bubble after {\forward} or {\backward} (for the corresponding micro-batch and layer), respectively.
    \item An {\inversion} work for $\mA_l$ or $\mB_l$ (for a layer) is assigned to a bubble after the {\curvature} work for $\mA_l$ or $\mB_l$ for all the micro-batches (for the corresponding layer), respectively.
    \item {\Precondition} work are assigned after {\backward} for all the layers in a stage and before the beginning of the next pipeline step.
\end{enumerate}
We first collect the profile of the CUDA kernel execution times of the standard work (i.e., {\forward} and {\backward}) during a step of a pipeline schedule followed by K-FAC work (i.e., {\curvature}, {\inversion}, and {\precondition}) on GPUs. 
Then we pick one work from the `queue' of all the K-FAC work and assign it to a bubble if its duration is shorter than the bubble duration (otherwise, subsequent bubbles are utilized) according to the rules above. 
We repeat this procedure until all the K-FAC work are assigned to bubbles.
Once all the K-FAC work are assigned (and the queue becomes empty), we finalize the (static) schedule and use it repeatedly until the training is completed. 
{\Curvature} and {\inversion} work often take a few steps (e.g., 2-3 steps) to complete (cf.$\sim$100-1000 steps in previous works), whereas {\precondition} is performed every step.
If the {\inverse} matrix of a layer is not ready, the one previously calculated for that layer is used for {\precondition}.

\autoref{fig:gpipe_1f1b_timeline} shows the profiling results (using NVIDIA's Nsight\footnote{\url{https://developer.nvidia.com/nsight-systems}}) of GPipe and 1F1B pipeline steps w/o and w/ automatically assigned K-FAC work.
Comparing the top and middle figures for GPipe and 1F1B, it can be seen that PipeFisher is making good use of the bubbles (\textbf{GPU utilization is increased from about 42\% to 89\%}), with {\precondition} work being the only major computational overhead.
In this setup, the {\curvature} and {\inverse} matrices are refreshed within a maximum of 2 steps (1 step for the second and third stages and 2 steps for the other stages).
%\htor{why is 42\% a good baseline?}

\subsection{Combination with data and {\inversion} parallelism} 
PipeFisher (and the automatic work assignments) can be combined with data and {\inversion} parallelism. 
The bottom figures in \autoref{fig:gpipe_1f1b_timeline} show the profiled timeline with the number of GPUs doubled (from 4 GPUs to 8 GPUs, 2 GPUs per stage.)
For the data parallelism, collective communication (\texttt{sync-grad} and \texttt{sync-curvature}) is performed between GPUs responsible for the same pipeline stage (e.g., GPU1 and GPU2 for stage1) to synchronize gradient and {\curvature}.  
Since the {\inversion} work (for 3 layers per stage) are split among 2 GPUs, the communication cost for {\curvature} synchronization is amortized. 
Thus, only gradient synchronization (as with distributed SGD and distributed Adam) is the main communication overhead.
%\footnote{In data- and {\inverse}-workload-parallel K-FAC, a reduce-scatter of gradients followed by an allgather of \textit{preconditioned} gradients is performed. These communication costs are the same as an all-reduce of gradients. \htor{I recommend removing all these dashes, it's allreduce and allgather in MPI, and MPI is the reference}}.

\begin{figure*}
    \centering
    \includegraphics[width=\textwidth]{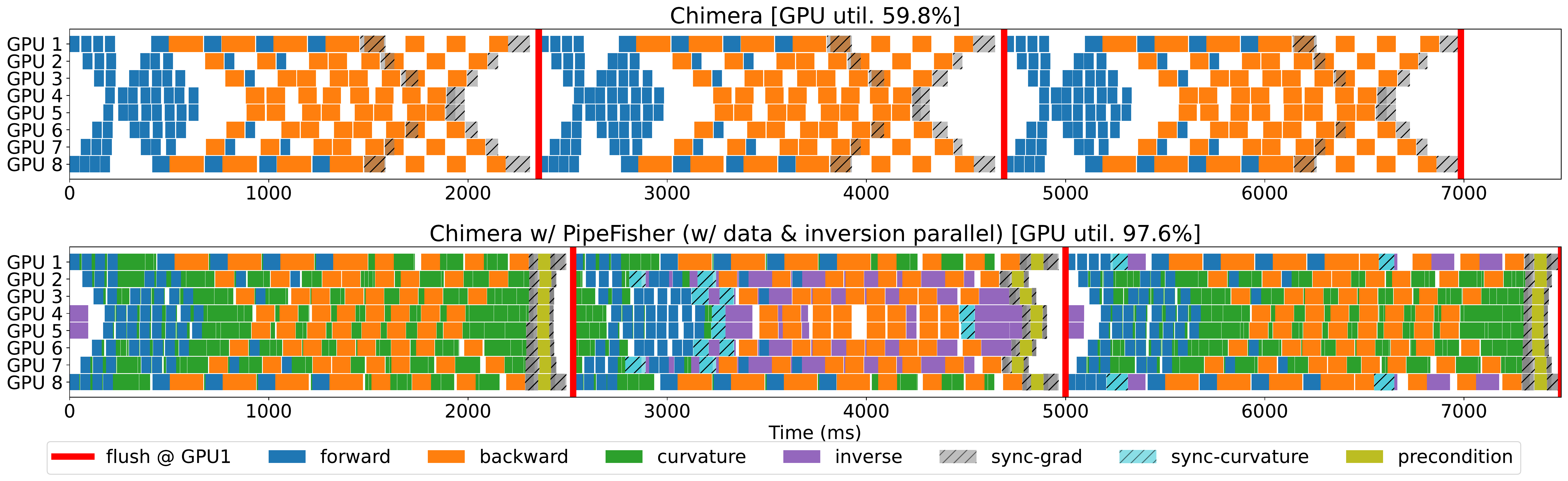}
    \caption{Profiled CUDA kernel execution times on NVIDIA P100 GPUs for 5-7th pipeline steps in Chimera \cite{li_chimera_2021} w/ Adam (top) and w/ PipeFisher w/ data and {\inversion} parallelism (bottom) in pretraining BERT-Large ($\nlayers=24$, ) w/ 8 stages (3 layers/stage), 8 GPUs, 8 micro-batches of size 32 per GPU per step, and sequence length 128.}
    \label{fig:chimera_timeline}
\end{figure*}

To better demonstrate the effectiveness of the automatic work assignments, we target the Chimera \cite{li_chimera_2021} pipeline schedule which is even more complex and more efficient than GPipe and 1F1B.
Chimera handles multiple pipelines simultaneously to make effective use of bubbles and efficiently increase throughput.
\autoref{fig:chimera_timeline} (top) shows the pipeline schedule in Chimera using two bidirectional pipelines (\textit{up pipeline} and \textit{down pipeline}). 
Since each GPU is responsible for two stages simultaneously, gradient synchronization (\texttt{sync-grad}) is performed for data parallelism between GPUs responsible for the same stage (e.g., GPU 1 and 8 for stage 1 and 8, GPU 4 and 5 for stage 4 and 5.)
\autoref{fig:chimera_timeline} (bottom) shows the results of applying PipeFisher (with data and {\inversion} parallelism) to Chimera; \textbf{PipeFisher increases GPU utilization from 59.8\% to 97.6\%}. 
With this setup, {\curvature} and {\inverse} matrices are refreshed in 4 steps for stages 1 and 8, and in 2 steps for the other stages.
%{\Precondition} workloads are the main computational overheads, but sometimes the assignments are too aggressive, slightly delaying the execution of {\forward} and {\backward} workloads.\htor{what's the purpose of this last sentence? Closing the section on a bad aftertaste?? :-( I'd remove.}

\subsection{Performance Modeling}
The number of pipeline steps required to complete the {\curvature} and {\inversion} work determines how often the curvature information is refreshed. 
%which can affects the convergence of second-order optimization.
Also, these work require additional memory consumption, which can limit the model size and micro-batch size.
To estimate the frequency of matrix updates and the total memory consumption, we create a performance model.
\autoref{tab:symbol} lists frequently used symbols.

\begin{table}[h]
    \centering
    \caption{Symbols}
    \begin{tabular}{ll}
    \toprule
       $\pldepth$  & The number of pipeline stages ({\it depth})  \\
%       $\plwidth$ & The number of replicated pipelines ({\it width}) for data parallelism \\
       $\nmicrobatches$ & \specialcelll{The number of micro-batches per device \\ within a training iteration} \\
       $\microbs$ & Micro-batch size \\
       $\minibs$ & Mini-batch size ($=\microbs*\nmicrobatches*\plwidth$) \\
%       $\memorydevice$ & The device memory \\
       $\memoryparam$ & \specialcelll{Memory consumption for the parameters of \\ one stage} \\
       $\memoryact$ & \specialcelll{Memory consumption for the activations of \\ one stage for one micro-batch} \\
       $\peakmemoryerr$ & \specialcelll{Peak memory consumption for the errors of \\ one stage for one micro-batch} \\
       $\countfwd$, $\countbwd$ & \specialcelll{The number of {\forward}, {\backward} passes on \\ the critical path in a pipeline step (when \\ $\nmicrobatches=\pldepth$, $\countfwd=\countbwd=2\pldepth-1$ for GPipe \\ and 1F1B (w/ pipeline flush) and $\countfwd=\pldepth$, \\ $\countbwd=2\pldepth-2$ for Chimera)} \\
    \bottomrule
    \end{tabular}
    \label{tab:symbol}
\end{table}

\begin{figure*}[t]
    \centering
    \begin{subfigure}{.9\linewidth}
        \includegraphics[width=\linewidth]{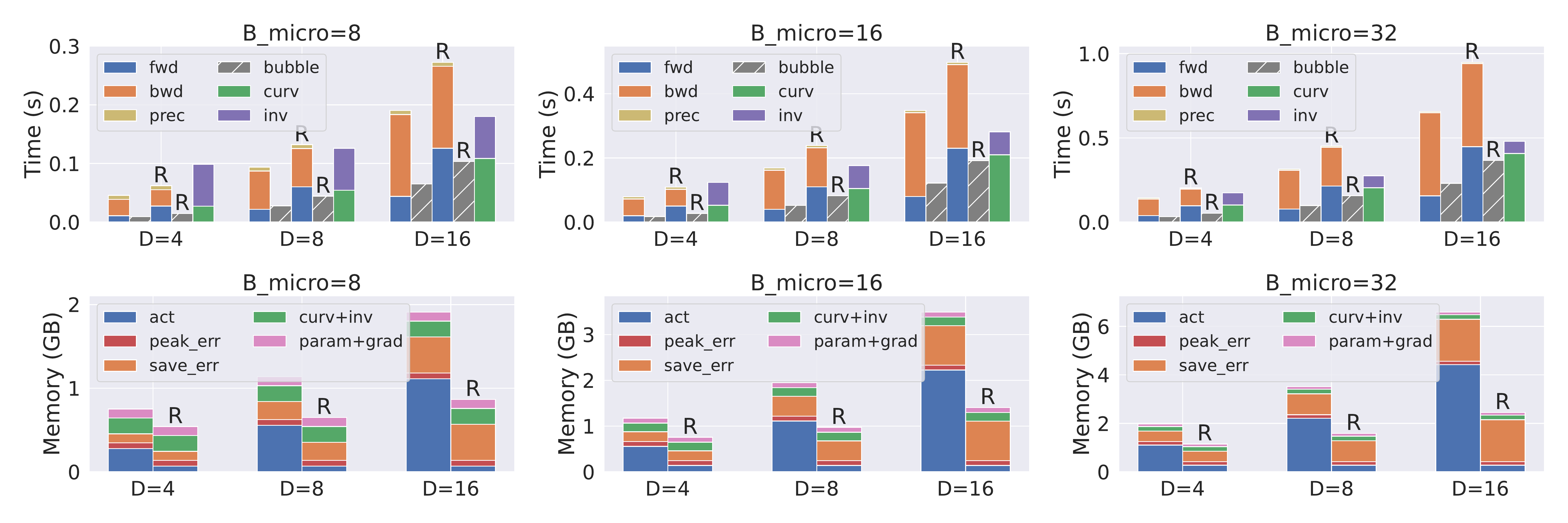}
        \caption{Modeled time per step and memory consumption}
    \end{subfigure}
    \begin{subfigure}{.9\linewidth}
        \includegraphics[width=\linewidth]{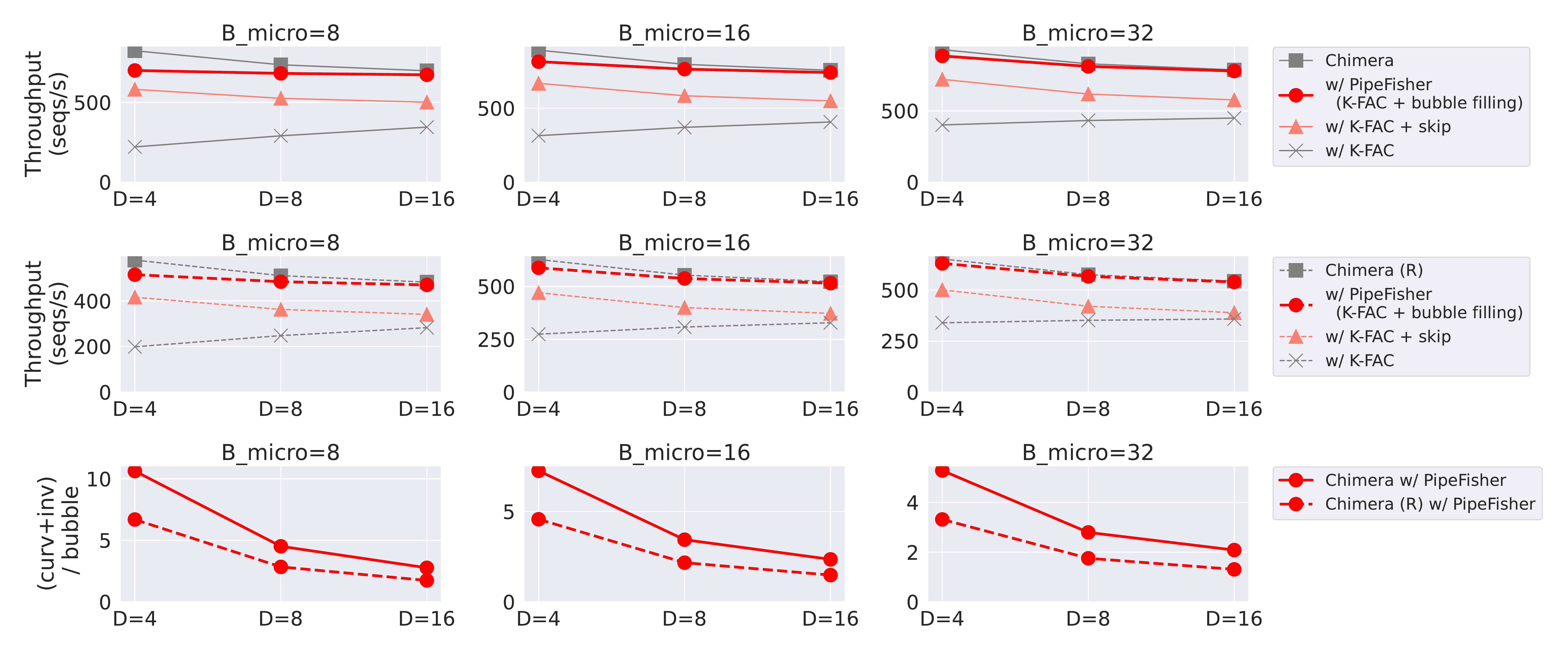}
        \caption{
        \change{Modeled throughput (sequences/s) and ({\curvature}+{\inversion})-bubble ratio}}
    \end{subfigure}
    \caption{Performance model for Chimera (w/ 2 pipelines) with $\pldepth$ BERT-Base blocks (one block per pipeline stage) with $\nmicrobatches=\pldepth$ on an NVIDIA P100. 
    \change{
    ``\textbf{w/ PipeFisher (K-FAC + bubble filling)}'': The pipeline bubbles are filled with the {\curvature} and {\inversion} work of K-FAC, and only the {\precondition} work is the computational overhead of PipeFisher over the vanilla Chimera.
    ``\textbf{w/ K-FAC + skip}'': Naive K-FAC execution (i.e., every K-FAC work is performed \textit{without} utilizing the bubbles) with the {\curvature} and {\inversion} work skipped every \texttt{(curv+inv)/bubble} iterations, i.e., the same frequency as PipeFisher.
    ``\textbf{w/ K-FAC}'': Naive K-FAC execution without skipping.
    }
    R indicates activation recomputation.
    }
    \label{fig:bert_base_perf_model_chimera}
\end{figure*}

For simplicity, P2P communication costs are ignored in the modeling since there are few gaps (i.e., latency for P2P communication) between {\forward}s and {\backward}s in the profile results in \autoref{fig:gpipe_1f1b_timeline} and \autoref{fig:chimera_timeline}.
We also ignore the cost of collective communication (i.e., \texttt{sync-grad} and \texttt{sync-curvature}) since our goal here is to model the duration of the bubbles and the size of the K-FAC work.
Assuming that all pipeline stages have the same size model partition, we put $\timefwd$ and $\timebwd$ as the computation time of {\forward} and {\backward} for one micro-batch, respectively.
Then one pipeline step time $\timepipe=\countfwd\timefwd+\countbwd\timebwd$ and the total bubble time $\timebubble=\timepipe-\nmicrobatches(\timefwd+\timebwd)$.
The worst case memory consumption (among all pipeline stages) is modeled as: $\memorypipe=2\frac{\pldepth\plwidth}{\#devices}\memoryparam+\nmicrobatches\memoryact+\peakmemoryerr$.
The time and memory overheads for the K-FAC work are:
\begin{align*}
\timekfac&=\underbrace{\nmicrobatches\timecurv+\timeinv}_{\mbox{fit into bubbles}}+\timeprec
\\
\mbox{and\,\,\,}
\memorykfac&=\memorycurv+\memoryinv+\nmicrobatches\savememoryerr
\,,
\end{align*}
where $\timecurv/\memorycurv$, $\timeinv/\memoryinv$, and $\timeprec$ represent the time/memory for {\curvature} (for one micro-batch), {\inversion}, and {\precondition} work for one stage, respectively ($\memorycurv=\memoryinv$.)
And $\savememoryerr$ is the memory cost to keep the errors $\err_l$ for calculating the Kronecker factors $\mB_l$ (for one micro-batch) (the memory cost to keep the activations $\act_l$ for $\mA_l$ is included in $\memorypipe$.)
We take microbenchmarks and measure the times and memories for different $\microbs$, $\pldepth$, pipeline methods (GPipe, 1F1B, or Chimera) and BERT models (Base or Large) on an NVIDIA P100 GPU.

\autoref{fig:bert_base_perf_model_chimera} shows the performance model for a pipeline stage of Chimera with a BERT-Base layer (assuming the BERT-Base model has $\pldepth$ layers in total).
Doubling the number of layers per pipeline stage doubles all times and memories.
Here we set $\nmicrobatches=\pldepth$, in which case the time, memory, and bubble ratio are the same in GPipe and 1F1B.

\change{\textbf{Computation time and throughput}:}
The top row of \autoref{fig:bert_base_perf_model_chimera} (a) shows the breakdown of computation time per step. 
There are five bars for each $(\microbs,\pldepth)$ combination, showing $\timepipe+\timeprec$, $\timebubble$, $\timepipe+\timeprec$ with activation recomputation (\textit{R}), $\timebubble$ with activation recomputation (\textit{R}), and $\timekfac-\timeprec$, where \textit{R} indicates the activation recomputation \cite{griewank_algorithm_2000} for saving memory consumption.
$\timepipe+\timeprec$ corresponds to the computation time per step of PipeFisher. 
Because $\timeprec$ is relatively small, the computational overhead of PipeFisher compared to the vanilla pipeline is small.
The top \change{and middle} row\change{s} of \autoref{fig:bert_base_perf_model_chimera} (b) compare the throughput of the vanilla pipeline and PipeFisher and show little difference between them.
As $\microbs$ and $\pldepth$ $(=\nmicrobatches)$ increase, $\timepipe$, $\timebubble$, and $\nmicrobatches\timecurv$ increase, while $\timeinv$ is constant regardless of $\microbs$ or $\pldepth$, so that $\nmicrobatches\timecurv+\timeinv<2\cdot\timebubble$ when $\microbs$ and $\pldepth$ is relatively large, and {\curvature} and {\inversion} work can be hidden within pipeline bubble in two pipeline iterations.
The bottom row of \autoref{fig:bert_base_perf_model_chimera} (b) shows the ratio of $\nmicrobatches\timecurv+\timeinv$ to $\timebubble$, suggesting the number of pipeline steps required for PipeFisher to refresh the curvature information.
\change{Through the effective use of bubbles, PipeFisher updates curvature information at a high frequency with a high throughput that cannot be achieved by simply skipping updates without utilizing the bubbles (``w/ PipeFisher'' vs. ``w/ K-FAC + skip'' in \autoref{fig:bert_base_perf_model_chimera} (b)).}

%\begin{figure*}
%    \centering
%    \begin{subfigure}{0.95\textwidth}
%        \includegraphics[width=\textwidth]{figures/bert_base_perf_model_gpipe_1f1b.pdf}
%        \caption{Modeled time per step and memory consumption}
%    \end{subfigure}
%    \begin{subfigure}{0.95\textwidth}
%        \includegraphics[width=\textwidth]{figures/bert_base_throughput_gpipe_1f1b.pdf}
%        \caption{Modeled throughput (sequences/s) and ({\curvature}+{\inversion})-bubble ratio}
%    \end{subfigure}
%    \caption{Performance model for $\pldepth$ BERT-Base blocks (one block per pipeline stage) with $\nmicrobatches=\pldepth$ in GPipe and 1F1B (with pipeline flush). R indicates activation recomputation.}
%    \label{fig:bert_base_perf_model_gpipe_1f1b}
%\end{figure*}

\change{\textbf{Memory consumption}:}
The bottom row of \autoref{fig:bert_base_perf_model_chimera} (a) shows the breakdown of memory consumption. 
There are two bars for each $(\microbs,\pldepth)$ combination, showing $\nmicrobatches\memoryact+\peakmemoryerr+\memoryparam+\memorykfac$ without/with \textit{R}.
$\nmicrobatches\memoryact$ and $\nmicrobatches\savememoryerr$ account for most of the memory consumption when $\microbs$ or $\pldepth$ $(=\nmicrobatches)$ is large, while $\memorycurv$ $(=\memoryinv)$ is constant.
Activation recomputation (\textit{R}) reduces throughput (due to the additional {\forward} work) and increases $\timebubble$, but at the same time significantly reduces memory consumption by $\memoryact$.
In this case, $\nmicrobatches\savememoryerr$, $\memorycurv$, and $\memoryinv$, i.e., $\memorykfac$, are the major bottlenecks.
As $\timebubble$ is increased by activation recomputation, curvature information is updated at a higher frequency.

\autoref{fig:bert_base_perf_model} and \autoref{fig:bert_large_perf_model} in \autoref{app:perf} summarize the performance models for BERT-Base and BERT-Large, respectively, with GPipe/1F1B or Chimera.
Chimera consistently achieves higher throughput than GPipe and 1F1B (due to the smaller $\timebubble$), but instead the curvature information is updated less frequently.
Therefore, the pipeline method can be selected based on the tradeoff between throughput and the frequency of extra information (i.e., curvature information for K-FAC) updates.

\begin{figure*}
    \centering
    \includegraphics[width=0.95\textwidth]{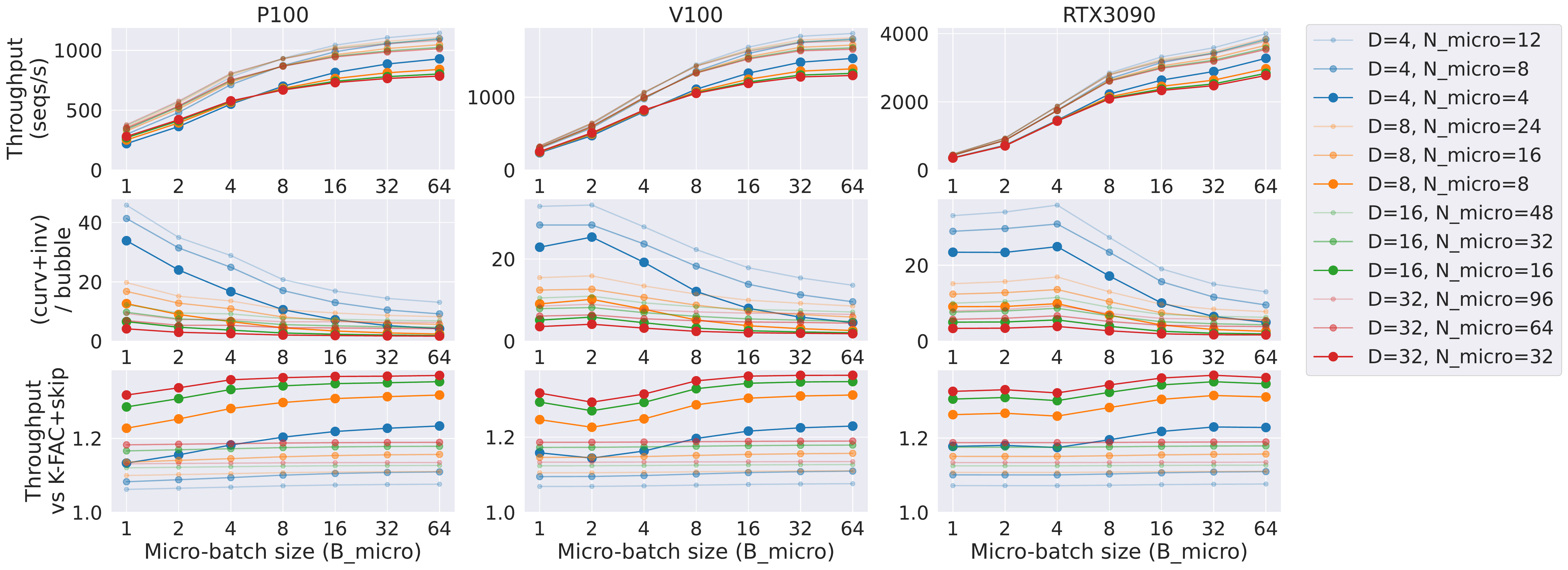}
    \caption{
    \change{Modeled throughput (sequences/s) and ({\curvature}+{\inversion})-bubble ratio of Chimera w/ $2$ pipelines w/ PipeFisher for $\pldepth$ {BERT-Base} blocks ($\pldepth\in\{4,8,16,32\}$, one block per pipeline stage) on an NVIDIA P100, V100, and RTX3090.
    ``\textbf{Throughput vs K-FAC+skip}'': Ratio of the throughput of PipeFisher to a naive K-FAC execution (i.e., every K-FAC work is performed \textit{without} utilizing the bubbles) with the {\curvature} and {\inversion} work skipped every \texttt{(curv+inv)/bubble} iterations, i.e., the same frequency as PipeFisher.
    }}
    \label{fig:bert_base_throughput2_chimera}
\end{figure*}

To better understand the scaling behavior of PipeFisher, we make the same observations (i.e., throughput and ({\curvature}+{\inversion})-bubble ratio) with various Transformer architectures (w/ different sequence lengths $\seqlen$), mini-/micro-batch sizes ($\minibs=\nmicrobatches\cdot\microbs$), and hardware (NVIDIA P100, V100, and RTX3090).
This time we will focus only on Chimera, which has fewer bubbles and achieves a higher throughput than GPipe and 1F1B.
\change{\autoref{fig:bert_base_throughput2_chimera} shows the results for BERT-Base.}
\change{Other} results are listed in \autoref{tab:figures}.
Below is a summary of observations:
\begin{itemize}
    \item Since the {\precondition} work (independent of $\minibs=\nmicrobatches\cdot\microbs$) is relatively small in all settings, little difference in throughput is observed between Chimera and Chimera w/ PipeFisher. Therefore, only the throughput of Chimera w/ PipeFisher is shown.
    \item As the micro-batch size $\microbs$ is increased, the ({\curvature}+{\inversion})-bubble ratio becomes smaller (i.e., easier to fit extra work to the bubbles) because the cost of the {\inversion} work is relatively small.
    \item Furthermore, as the pipeline depth $\pldepth$ increases, the ratio goes down because the bubble increases.
    \item On the other hand, as the number of micro-batches $\nmicrobatches$ is increased, the ratio increases because the bubbles become smaller.
    \item Transformers with longer sequence lengths $\seqlen$ have larger bubbles and smaller ratios. 
    This is because the total number of tokens ($\seqlen\cdot\nmicrobatches\cdot\microbs$) linearly increases the {\forward}, {\backward}, and {\curvature} work, while {\inversion} work is independent of it.
    \item The change in the ratio in different hardware depends on the Transformer architecture (i.e., a faster GPU increases, decreases, or does not change the ratio).
    \item In most cases the ratio is in the range of 2-10, except when the micro-batch size $\microbs$ is particularly small (e.g., 1,2) and the number of micro-batches $\nmicrobatches$ is large (e.g., $\nmicrobatches=3\pldepth$). This suggests that curvature information is updated at a high frequency.
    \item \change{PipeFisher provides up to about $1.4\times$ speedup versus naive K-FAC execution with {\curvature}-/{\inversion}-skipping (``K-FAC+skip'') when $\nmicrobatches=\pldepth$ and $\microbs$ is large (64). On the other hand, when the number of micro-batches is large (e.g., $\nmicrobatches=3\pldepth$) or $\microbs$ is small, speedup by PipeFisher is limited to about $1.1\times$.}
\end{itemize}

\section{Language Modeling}
\label{sec:experiments}
We apply PipeFisher to the pretraining of BERT-Base and -Large models \cite{devlin_bert_2019} on the English Wikipedia \cite{wikipedia_wikimedia_nodate} (see Appendix for information on preparing the dataset.)
The task is to minimize the sum of the masked language modeling loss (classification with vocabulary size 30,522) and next sentence prediction loss (binary classification).
BERT pretraining consists of two phases, where the maximum sequence lengths are 128 and 512, respectively.
%\htor{what else differs? Learning rate? etc.}
The learning rate, mini-batch size and number of steps for each phase depends on the implementation, but the number of steps in Phase 1 often accounts for 80-90\% of the total (90\% in the original work \cite{devlin_bert_2019}).
We use NVLAMB, NVIDIA's implementation of the LAMB optimizer \cite{you_large_2020}, as the baseline optimizer to be compared with K-FAC (with PipeFisher).
As the full pretraining of BERT requires a huge amount of energy and CO$_2$ overheads, we only discuss the training time in Phase 1.
Following Pauloski et al. (2022) \cite{pauloski_deep_2022}, we apply K-FAC to all fully-connected layers except for the final classification head, where $\dout_\nlayers=30,522$ (vocabulary size) and the Kronecker factor $\mB_\nlayers$ will be too large to construct/invert, and we use NVLAMB for the rest of layers. 
Hereafter, for simplicity, we refer to this as K-FAC.

\begin{figure*}
    \centering
    \includegraphics[width=.85\textwidth]{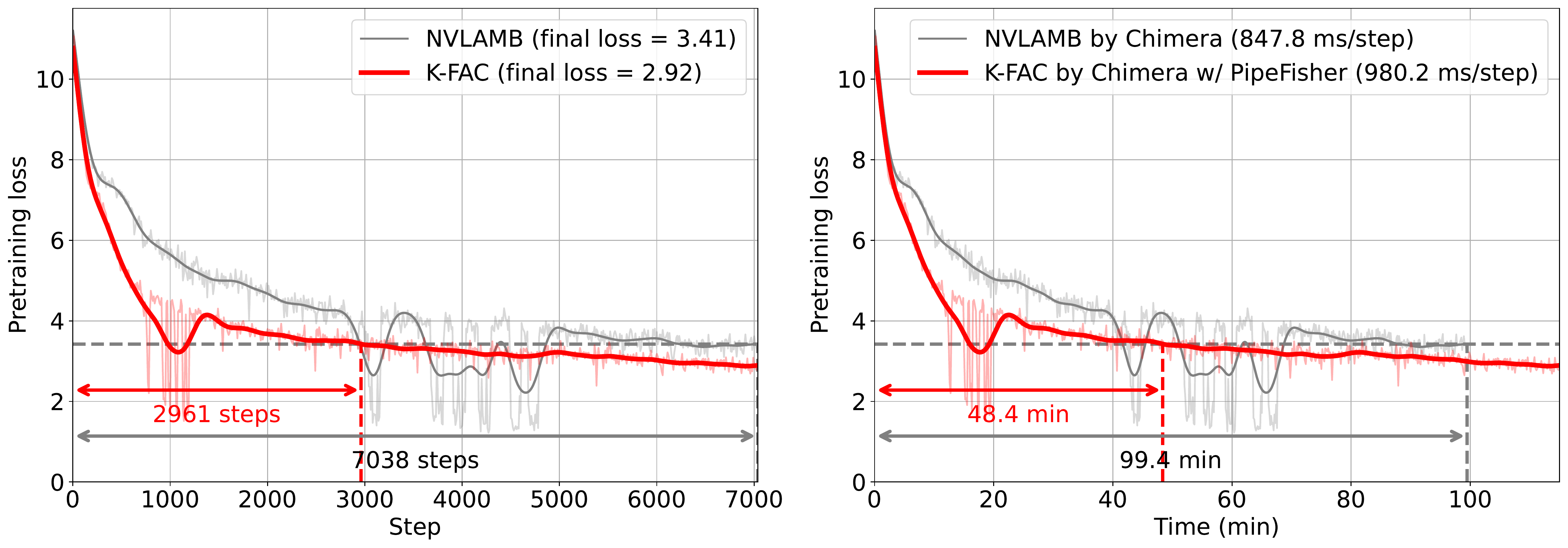}
    \caption{\textbf{Left}: Pretraining BERT-Base on the English Wikipedia (Phase 1: mini-batch size 8K, sequence length 128 for 7,038 steps) using NVLAMB and K-FAC. 
    \textbf{Right}: Same training curves as on the left applying time per step measured on 256 P100 GPUs using Chimera for NVLAMB (GPU util.: 75.9\%) and Chimera \textbf{with PipeFisher} for K-FAC (\textbf{GPU util.: 93.2\%}) (with 4 pipeline stages). With this setup, PipeFisher updates the {\inverse} every 5 to 10 steps.
    Each curve is smoothed by applying signal filters using SciPy v1.8.0 (\texttt{y\_smooth = signal.filtfilt(*signal.butter(3, 0.05), y)}). 
    The number of steps and time for K-FAC to reach NVLAMB's final loss (3.41) are calculated for the smooth curves (ignoring  large fluctuations around the 1,000th step.)
    }
    \label{fig:bert-base}
\end{figure*}

\textbf{BERT-Base}: We first train BERT-Base using NVLAMB and K-FAC. 
Our training code is based on NVIDIA's PyTorch-based codebase for BERT pretraining\footnote{\url{https://github.com/NVIDIA/DeepLearningExamples}} and we use the same training hyperparamers for NVLAMB --- mini-batch size 8,192, weight decay 0.01, base learning rate $6\cdot 10^{-3}$, training steps 7,038, and linear learning rate warming up steps 2,000. 
For K-FAC, the same hyperparameters are used except that the number of learning rate warming up steps is reduced to 600, resulting in larger learning rates than NVLAMB until the 2,000th step (see Appendix for more information on training settings.)
We observed that during the first 1,000-2,000 steps, K-FAC benefits from a more aggressive learning rate, whereas training diverges when the same learning rate is applied to NVLAMB.
\autoref{fig:bert-base} shows the pretraining loss versus the number of steps and training time.
K-FAC significantly improves the convergence and reaches NVLAMB's final loss (3.41) in 2,961 steps (42.0\% of 7038 steps).
For measuring the wall-clock time, we run NVLAMB by Chimera and K-FAC by Chimera with PipeFisher on 256 GPUs with 4 pipeline stages (thus 64 model copies) and 4 micro-batches of size 32 per optimization step ($4\times 32\times 64=8,192$.)
As the {\precondition} work is the only major computational overhead, PipeFisher retains the improved convergence by K-FAC and reaches NVLAMB's final loss in \textbf{48.4 minutes (48.7\% of 99.4 minutes)} while \textbf{improving the GPU utilization from 75.9\% to 93.2\%}.

\textbf{BERT-Large}: Next, we target BERT-Large model. 
Since pretraining BERT-Large is resource-intensive, we rely on the results of Pauloski et al. (2022) \cite{pauloski_deep_2022} for the number of training steps by NVLAMB and (data- and {\inversion}-parallel) K-FAC (with {\inverse} refreshed once every 50 steps) and the SQuAD v1.1 F1 score after fine tuning.
In addition to this, their Phase 1 results use a mini-batch of size 64K, so setting the micro-batch size to 32 (maximum number of powers of 2 that can be placed on a P100 GPU) would require a 2K GPUs, which requires a huge computing budget. %\htor{that is not true and also sounds very bad}
So, we instead apply the time per step with Chimera w/ and w/o PipeFisher measured on 8 GPUs as in \autoref{fig:chimera_timeline} to simulate the training time, i.e., ignoring the increase in communication costs when scaling from 8 GPUs to 2K GPUs. %\htor{that now sounds super-bad!}
The results are summarized in \autoref{tab:bert-large}.
As the computational overhead per step with PipeFisher is only $\sim$6.5\%, the Phase 1 training time is \textbf{reduced from 275.1 to  208.3 minutes (75.7\%)} (in the simulation) by \textbf{improving the GPU utilization from 59.8\% to 97.6\%} and taking advantage of the convergence improved by K-FAC.

\begin{table*}[t]
    \centering
    \caption{Pretraining BERT-Large on the English Wikipedia (Phase 1: mini-batch size 64K, Phase 2: mini-batch size 32K) using NVLAMB and K-FAC. 
    We use the results from \cite{pauloski_deep_2022} for the number of steps in Phase 1 and Phase 2 and the SQuAD v1.1 F1 score after fine tuning.
    * We apply time per step measured on 8 P100 GPUs using Chimera for NVLAMB (GPU util.: 59.8\%) and Chimera \textbf{with PipeFisher} for K-FAC (\textbf{GPU util.: 97.6\%}) (same setting as \autoref{fig:chimera_timeline}) to simulate the training time.}
    \begin{tabular}{ccccccc}
    \toprule
    \multirow{2}[2]{*}{Optimizer} & \multirow{2}[2]{*}{Pipeline scheme} & \multicolumn{3}{c}{Phase 1} & Phase 2 & \multirow{2}[2]{*}{F1} \\ 
    \cmidrule(lr){3-5} \cmidrule(lr){6-6}
    & & Steps & Time/step$^*$ & Time$^*$ & Steps & \\ 
    \midrule
    NVLAMB & Chimera & 7038 & 2345.6 ms & 275.1 min & 1563 & 90.1\% \\
    K-FAC & Chimera \textbf{w/ PipeFisher} & 5000 & 2499.5 ms & 208.3 min & 1563 & 90.15\% \\ 
    \bottomrule \\
    \end{tabular}
    \label{tab:bert-large}
\end{table*}
\section{Discussion and Conclusion}
\label{sec:discussion}

\textbf{PipeFisher for non-Transformer architectures}:
PipeFisher is applicable to any neural architecture that can be pipelined.
As a Transformer model is composed of multiple encoder/decoder layers of the same size (except for embedding layers and task-specific heads), it is easy to distribute the work equally among pipeline stages, making it a particularly good match for pipelining and PipeFisher.
On the other hand, other architectures, such as convolutional neural networks, often have different numbers of neurons/channels and feature map sizes at each layer, so it is more challenging to apply pipelining and divide the work evenly.
In particular, the computational cost of {\inversion} work is proportional to the cube of the matrix size, which can easily cause load imbalance. %\htor{spell check!}

\textbf{Extra work for other types of algorithms}:
The application of the idea of ``assigning extra work to bubbles in pipeline for auxiliary benefits'' is not limited to K-FAC. 
For example, pipelining the work of Shampoo optimizer \cite{gupta_shampoo_2018}, which also accelerates training Transformers \cite{anil_scalable_2021} and requires Kronecker-factored matrices of the same size as the K-FAC (for fully-connected layers), is a natural extension of the PipeFisher.
Since the Shampoo optimizer requires an eigenvalue decomposition, which is computationally more expensive than an {\inversion}, for each matrix, a method that divides the work for a single matrix into multiple pieces would be necessary for an efficient bubble utilization.
Another example, other than training acceleration, is the improvement of generalization performance through Sharpness-Aware Minimization (SAM) \cite{foret_sharpness-aware_2021}.
SAM requires an additional {\forward} and {\backward} for every training step to estimate the loss sharpness \cite{hochreiter_flat_1997,keskar_large-batch_2017}, so it contains twice the work of regular SGD and has the potential to double the accelerator utilization.

\textbf{Limitations}: 
To avoid the enormous energy and CO$_2$ overheads of pretraining LLMs, we do not conduct end-to-end time measurements. 
Instead, we simulate the time by multiplying the measured time per step by the total number of steps. 
%\htor{I'd argue why those are accurate and we didn't do it to save CO2}
For this reason, although the ability to update curvature information frequently is one advantage of PipeFisher over existing distributed K-FAC approaches (see the first paragraph of Section~\ref{sec:pipefisher}), this study does not analyze its effect on the convergence.  
%\htor{can we not cite something here and say we exclude it consciously?}
%Yet, we believe that the ability to experiment at high frequencies in large settings is valuable for future analysis.
Because of the limited scope of the target model/task and the hyperparameter search (we only changed the number of learning rate warming up steps), this study does not prove the general advantages of K-FAC over other optimizers.
Yet, PipeFisher enables a cheaper hyper-paramerer search (see Appendix C.2).
%or second-order methods over first-order methods.

\textbf{Conclusion}: 
In this study, we demonstrate how much free time exists in pipeline-parallel training, one important component of LLM training, and how large work (computation and communication) can be packed into it by careful profiling and visualization.
We propose \textit{PipeFisher}, which automatically assigns the work of K-FAC \citep{martens_optimizing_2015}, an optimization method based on the Fisher information matrix, to bubbles in any pipeline scheme, and show that it considerably increases GPU utilization and reduces (simulated) Phase 1 pretraining time for BERT-Base and -Large to 50-75\%.
The improved convergence by K-FAC is one example of the benefits we can gain from the extra work.
We believe that our study will inspire other ``filling bubbles'' approaches that efficiently improve large-scale training.

% Acknowledgements should only appear in the accepted version.
%\section*{Acknowledgements}

\bibliographystyle{mlsys2023}
\bibliography{references}

\appendix
\section{Performance Analyses}
\label{app:perf}
%We create a performance model to estimate the frequency of {\curvature} and {\inverse} matrix updates and the memory overhead in PipeFisher.
%\autoref{tab:symbol} lists frequently used symbols.

\subsection{Comparisons on various Transformers, mini-/micro-batch sizes, and hardware}
\autoref{tab:figures} lists all the performance model figures and the corresponding Transformer configurations.

\subsection{PipeFisher for larger Transformers}
As $\embdim$ and $\ffdim$ in \autoref{tab:figures} correspond to the sizes of the {\curvature} and {\inverse} matrices (i.e., $\din_l,\dout_l$), if these are increased (e.g., 16,384), the matrices are too large to be placed in GPU memory.
For this reason, we limit our observations to the "Base" and "Large" models.
For even larger Transformer models, a possible strategy would be to approximate each {\curvature} matrix as a block diagonal matrix, thereby reducing memory and {\curvature}+{\inversion} work costs (this has already been incorporated in Shampoo for BERT pre-training \cite{anil_scalable_2021}, which requires matrix computations of the same size as K-FAC).
If $\embdim$ and $\ffdim$ are multiplied by $K$ and each matrix is approximated by a $K$-block diagonal matrix (e.g., an {\inversion} work of size 16,384 will be split into four {\inversion} work of size 4,096 when $K=4$), then the computation for all work ({\forward}, {\backward}, {\curvature}, {\inversion}, and  {\precondition}) and bubble times are $K$ times longer. 
Therefore, the ({\curvature}+{\inversion})-bubble ratio (i.e., how many pipeline iterations are required to refresh the curvature information) will match the value before scaling by $K$, and a similar work assignment can be used.

\begin{table*}[]
    \centering
    \caption{
    List of performance model figures for each Transformer architecture. 
    \texttt{Block class}: corresponding Python class that defines the Transformer block (a multi-head self-attention followed by a feed forward layer) in Hugging Face's \texttt{transformers}.
    $\embdim$: dimensionality of the encoder layer, $\ffdim$: dimensionality of the intermediate feed forward layer, $\nheads$: number of the attention heads, $\seqlen$: sequence length.}
    \begin{tabular}{lllcccc}
    \toprule
    \multirow{2}[2]{*}{Figure} & \multirow{2}[2]{*}{Architecture} & \multirow{2}[2]{*}{Block class} & \multicolumn{4}{c}{Configuration} \\
    \cmidrule{4-7}
    &&& $\embdim$ & $\ffdim$ & $\nheads$ & $\seqlen$ \\
    \midrule
    \autoref{fig:bert_base_perf_model}, \autoref{fig:bert_base_perf_model2} & BERT-Base & \texttt{BertLayer} & 768 & 3072 & 12 & 128 \\
    \autoref{fig:bert_large_perf_model}, \autoref{fig:bert_large_perf_model2} & BERT-Large & \texttt{BertLayer} & 1024 & 4096 & 16 & 128 \\
    \autoref{fig:t5_base_perf_model2} & T5-Base & \texttt{T5Block} & 768 & 3072 & 12 & 512 \\
    \autoref{fig:t5_large_perf_model2} & T5-Large & \texttt{T5Block} & 1024 & 4096 & 16 & 512 \\
    \autoref{fig:opt_125m_perf_model2} & OPT-125M (Base) & \texttt{OPTDecoderLayer} & 768 & 3072 & 12 & 2048 \\
    \autoref{fig:opt_350m_perf_model2} & OPT-350M (Large) & \texttt{OPTDecoderLayer} & 1024 & 4096 & 16 & 2048 \\
    \bottomrule
    \end{tabular}
    \label{tab:figures}
\end{table*}

\section{Experimental Settings}
\subsection{Training data}
To prepare the 14 GB English Wikipedia \cite{wikipedia_wikimedia_nodate}, we follow the data preparation instruction provided by Microsoft\footnote{\url{https://github.com/microsoft/AzureML-BERT/blob/master/docs/dataprep.md}}.
As described in the License information\footnote{\url{https://dumps.wikimedia.org/legal.html}}, ``all original textual content is licensed under the GNU Free Documentation License (GFDL) and the Creative Commons Attribution-Share-Alike 3.0 License.''
The Term of Use\footnote{\url{https://foundation.wikimedia.org/wiki/Terms_of_Use/en}} says ``you may encounter material that you find offensive, erroneous, misleading, mislabeled, or otherwise objectionable.''

\subsection{Training settings}
\begin{figure}
    \centering
    \includegraphics[width=\linewidth]{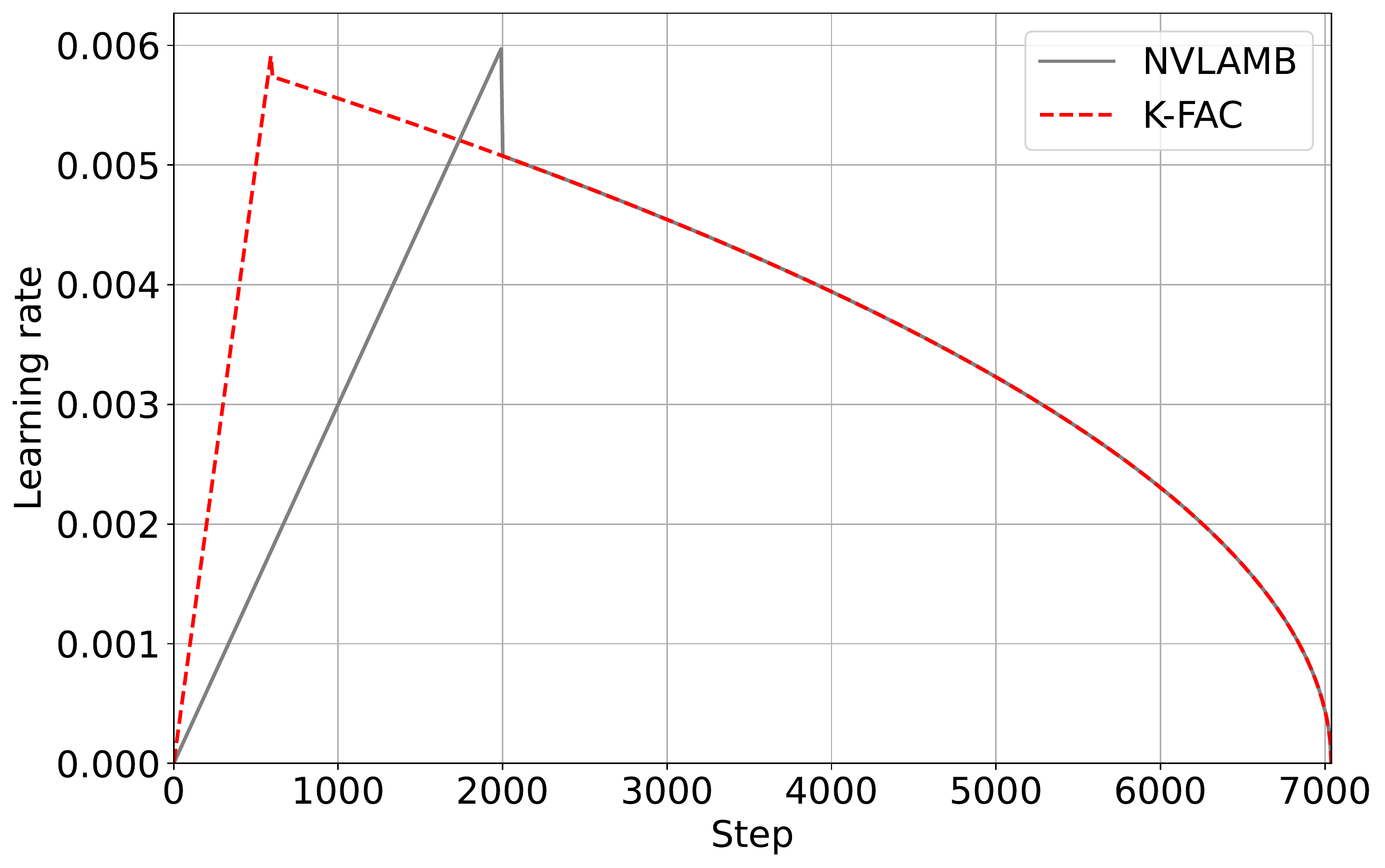}
    \caption{Learning rate schedule in Phase 1 pretraining of BERT-Base}
    \label{fig:bert_base_lr}
\end{figure}

We pretrain BERT-Base \cite{devlin_bert_2019} on the English Wikipedia (Phase 1 only) by NVLAMB and K-FAC.
For NVLAMB, we set mini-batch size 8,192, max sequence length 128, weight decay 0.01, base learning rate $6\cdot 10^{-3}$, total training steps 7,038, and linear learning rate warming up steps 2,000.
The learning rate at the $t$-th step after warm-up is determined by the polynomial decay: $\eta_t=\mathrm{base\_lr}\times(1-t/\mathrm{total\_steps})^{0.5}$.
For K-FAC, the same hyperparameters are used except that the number of learning rate warming up steps is reduced to 600, resulting in larger learning rates than NVLAMB until the 2,000th step.
The pretraining loss versus the number steps is shown in Figure~5 (left).
\autoref{fig:bert_base_lr} shows the learning rate schedule.

Setting the micro-batch size to 32 (maximum number of powers of 2 that can be placed on a P100 GPU) would require 256 GPUs to run training with mini-batch size 8,192.
However, to reduce total GPU hours and energy and CO$_2$ overheads, we simulate this training by using 32 GPUs and accumulating the micro-batch gradient over 8 steps before updating parameters ($32\times 32\times 8 =8,192$.)\footnote{NVLAMB on 32 GPUs takes 3.74 seconds per parameter update (with a mini-batch of size 8,192) while 128 GPUs takes 1.23 seconds. Hence the speedup (3.04x) is not linear to the number of GPUs.}

In addition to this, the training is done using simple data parallelism \textbf{without pipelines} for reducing GPU hours.
This is because the entire BERT-Base model fits into the P100 GPU device memory (16 GB), and in this case data parallelism without any model partitioning saves the most GPU hours on 32 GPUs in the GPU cluster we use (although it increases the communication cost of the allreduce of gradients for data parallelism.)
While the target of model partitioning is a model that is too large to fit in the memory of a single device, our study simulates the effects of pipelining with relatively small Transformers (i.e. BERT-Base and -Large) compared to today's GPU memory limitations.
Yet, the same techniques, discussions, and benefits of pipelining (and PipeFisher) described in our study are applicable to even larger Transformers.

The choice of the parallel training strategy does not affect the convergence of NVLAMB as long as the micro-batch gradients are synchronized\footnote{We use the fp32 precision for every quantity (parameters, gradients, optimization state) in training, so we assume that the effect of the numerical precision is negligible.}, which is the case in all of our experiments.
For K-FAC, we use data- and {\inversion}-parallel K-FAC (Figure~2 (ii,b)), and the {\curvature} and {\inverse} matrices are refreshed once in 10 steps.
We assume this does not affect the convergence by PipeFisher because it refreshes the matrices more frequently (once in 5-10 steps) in this BERT-Base setup as described in Figure~5.

\subsection{Computational resources}
%We used the Piz Daint supercomputer with NVIDIA P100 GPUs for all experiments.
We use a GPU cluster\footnote{the cluster name is not shown for anonymity.} with NVIDIA P100 GPUs for all the experiments (except for Figure \ref{fig:bert_base_perf_model2},\ref{fig:bert_large_perf_model2},\ref{fig:t5_base_perf_model2},\ref{fig:t5_large_perf_model2},\ref{fig:opt_125m_perf_model2}, and \ref{fig:opt_350m_perf_model2} where we use an NVIDIA V100 and a RTX3090 for micro benchmarks).
For Phase 1 pretraining of BERT-Base, NVLAMB takes $\sim$7.4 hours while K-FAC takes $\sim$8.4 hours on 32 GPUs.
To measure the time per step of PipeFisher, we only need to run about 10 steps of training on 4 (for Figure~3), 8 (for Figure~3 and 4), or 256 GPUs (for Figure~5), so the execution time is about 1-2 minutes, which is negligible compared to the training costs.

\subsection{GPU utilization}
We profile the pipeline steps by NVIDIA's Nsight.
We extract the CUDA activities (\texttt{CUPTI\_ACTIVITY\_KIND\_KERNEL}) occurring within a work (either {\forward}, {\backward}, {\curvature}, {\inversion}, or {\precondition}) from the profile results, and their start and end intervals are colored with the corresponding color in Figure 3 and 4.
Therefore, the percentage of colored areas in each figure corresponds to the percentage of time that some kernel is being executed on the GPU, which we display as ``GPU utilization''.

\section{Additional Discussion}

\subsection{Asynchronous pipeline methods} 
A \textit{synchronous} pipeline method waits until the gradient calculations for all micro-batches in one mini-batch are completed at all pipeline stages before updating model parameters (\textit{pipeline flush}) and starting the next pipeline. 
Hence, the pipeline flush makes most accelerator devices idle and creates the pipeline bubble.
In \textit{asynchronous} pipeline methods (e.g., PipeDream \cite{narayanan_pipedream_2019}, PipeDream-2BW \cite{narayanan_memory-efficient_2021}), on the other hand, no pipeline flush is performed and a different version of the model parameters (from 1 up to $\pldepth$ (the pipeline depth) steps old) are used at each stage to calculate the gradient.
Therefore, pipeline bubbles are almost non-existent in asynchronous pipelines, but may reduce convergence in the gradient-based optimizer.
%In this paper, we refer 1F1B (proposed by Narayanan et el. (2019) \cite{narayanan_pipedream_2019}) as a synchronous pipeline method, although 1F1B is actually the scheduling scheme adopted by PipeDream, an asynchronous pipeline method. 
%What we call ``1F1B'' is a combination of the 1F1B scheduling with pipeline flush, which is equivalent to DAPPLE \cite{fan_dapple_2021} (2021) and PipeDream-Flush \cite{narayanan_memory-efficient_2021} (2021).

We propose PipeFisher as an extension to the synchronous pipeline methods for gaining an auxiliary benefit, i.e., improved convergence by K-FAC, in LLM training by ``filling bubbles'' with K-FAC work (i.e., {\curvature}, {\inversion}, and {\precondition} work.)
The model parameters $\params_t$ at the $t$-th step are updated by the fresh gradients $\vg_t$ preconditioned by the stale curvature information: $\params_{t+1}=\params_{t}-\eta\empfisher_{t-n}^{-1}\vg_{t}$, where $n$ represents the number of additional steps (from 1 to $\sim$10, depending on the choice of the synchronous pipeline method, micro-batch size, and so on) taken to refresh the curvature information, and $\eta$ is the learning rate.
We can also see an asynchronous pipeline method as a ``filling bubbles'' approach --- the bubbles are filled by the gradient calculation (i.e., {\forward} and {\backward} work) with the stale model parameters, resulting in a higher throughput (number of tokens processed per unit time). 
The model parameters are then updated by the stale gradients: $\params_{t+1}=\params_{t}-\eta\vg_{t-m}$, where $m$ represents the number of the steps (from 1 up to $\pldepth$ steps, depending on the choice of the asynchronous pipeline method) to refresh the gradients.

\subsection{Hyper-parameters for K-FAC}
Compared to Adam, the only additional hyper-parameter of K-FAC is the frequency of matrix updates (i.e., {\curvature} work and {\inversion} work). 
With PipeFisher, the update frequency is no longer a hyperparameter, but is determined by network structure, number of pipeline stages, micro-batch size, and hardware. 
Therefore, there is no need to tune the frequency to make a trade-off between training time and convergence, and the achievable frequency by PipeFisher is much higher than previously feasible. 
As PipeFisher is the implementation of K-FAC to be performed more accurately (i.e., a higher frequency of matrix update) and cheaply (i.e., no extra communication, less memory consumption, no tuning of the frequency is required), it enables a cheaper hyper-paramerer search.

\begin{figure*}
    \centering
    \begin{subfigure}{.8\linewidth}
        \includegraphics[width=\linewidth]{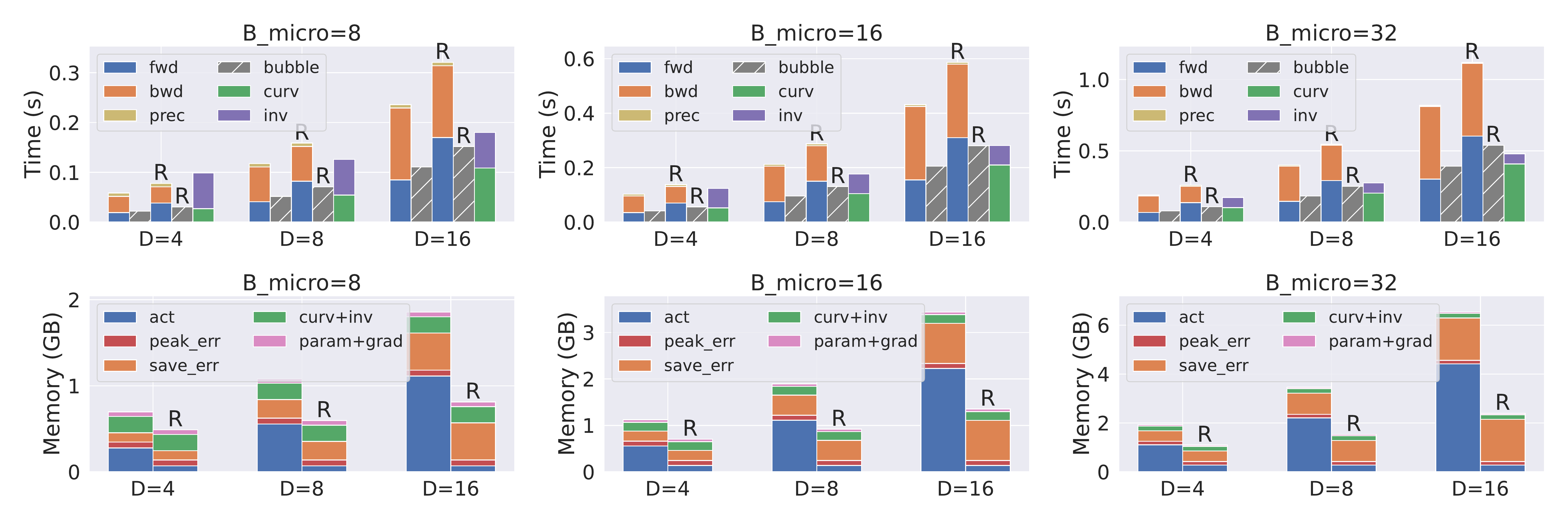}
        \caption{Modeled time per step and memory consumption of GPipe and 1F1B (w/ pipeline flush)}
    \end{subfigure}
    \begin{subfigure}{.8\linewidth}
        \includegraphics[width=\linewidth]{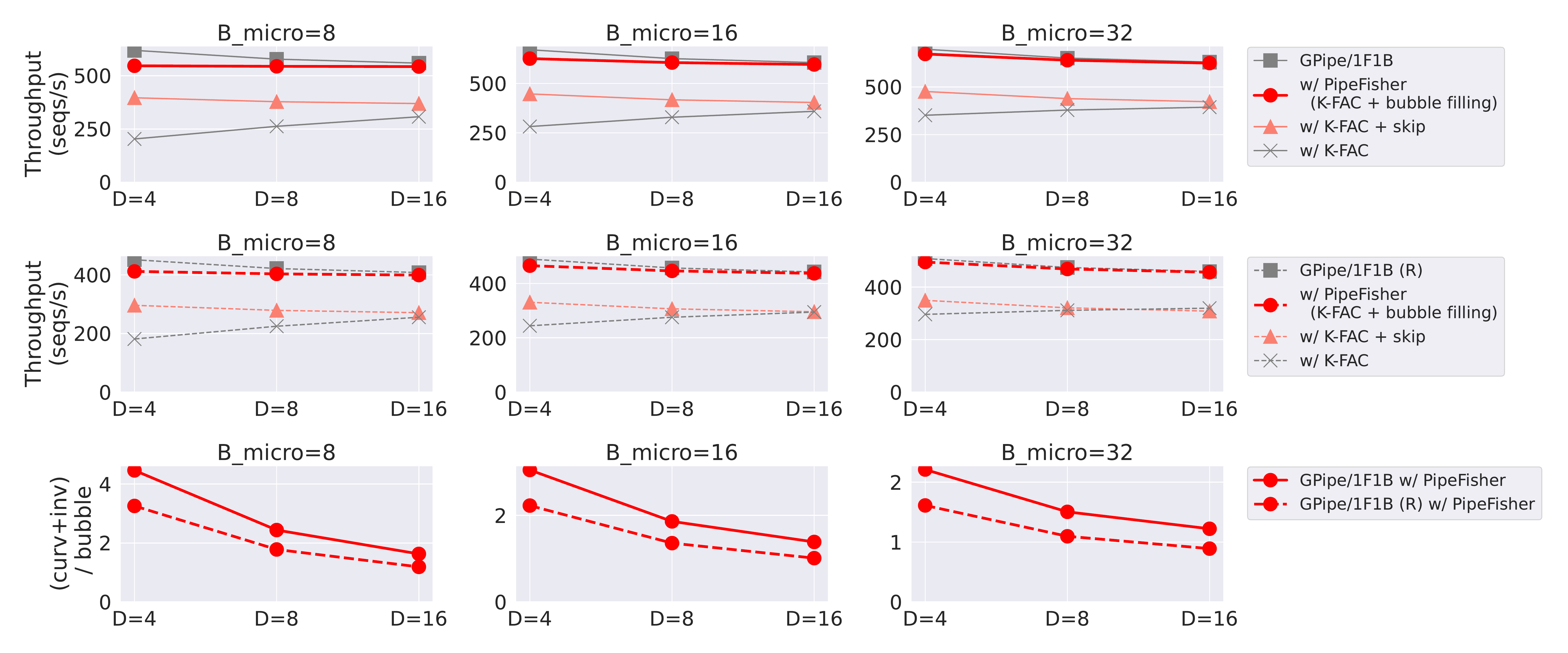}
        \caption{
        \change{Modeled throughput (sequences/s) and ({\curvature}+{\inversion})-bubble ratio of GPipe and 1F1B (w/ pipeline flush)}}
    \end{subfigure}
    \begin{subfigure}{.8\linewidth}
        \includegraphics[width=\linewidth]{figures/bert_base_perf_model_chimera.pdf}
        \caption{Modeled time per step and memory consumption of Chimera w/ $2$ pipelines}
    \end{subfigure}
    \begin{subfigure}{.8\linewidth}
        \includegraphics[width=\linewidth]{figures/bert-base_throughput_chimera.pdf}
        \caption{\change{Modeled throughput (sequences/s) and ({\curvature}+{\inversion})-bubble ratio of Chimera w/ $2$ pipelines}}
    \end{subfigure}
    \caption{Performance model for $\pldepth$ BERT-Base blocks (one block per pipeline stage) with $\nmicrobatches=\pldepth$ on an NVIDIA P100. R indicates activation recomputation.}
    \label{fig:bert_base_perf_model}
\end{figure*}

\begin{figure*}[h]
    \centering
    \begin{subfigure}{.8\linewidth}
        \includegraphics[width=\textwidth]{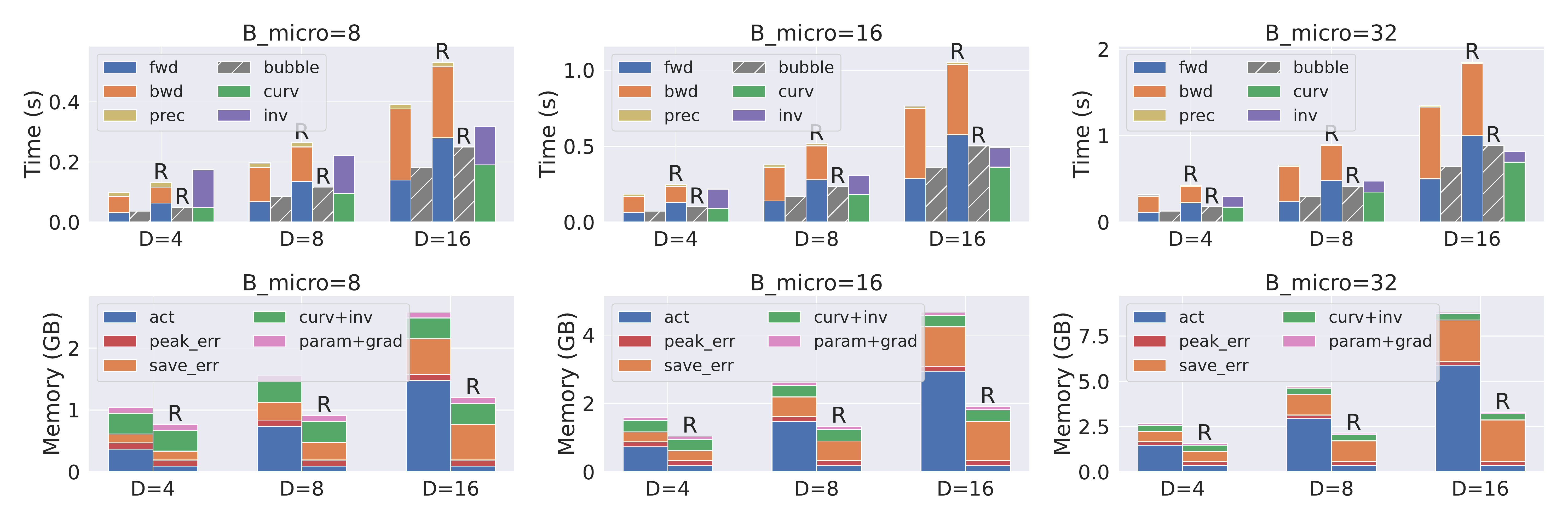}
        \caption{Modeled time per step and memory consumption of GPipe and 1F1B (w/ pipeline flush)}
    \end{subfigure}
    \begin{subfigure}{.8\linewidth}
        \includegraphics[width=\textwidth]{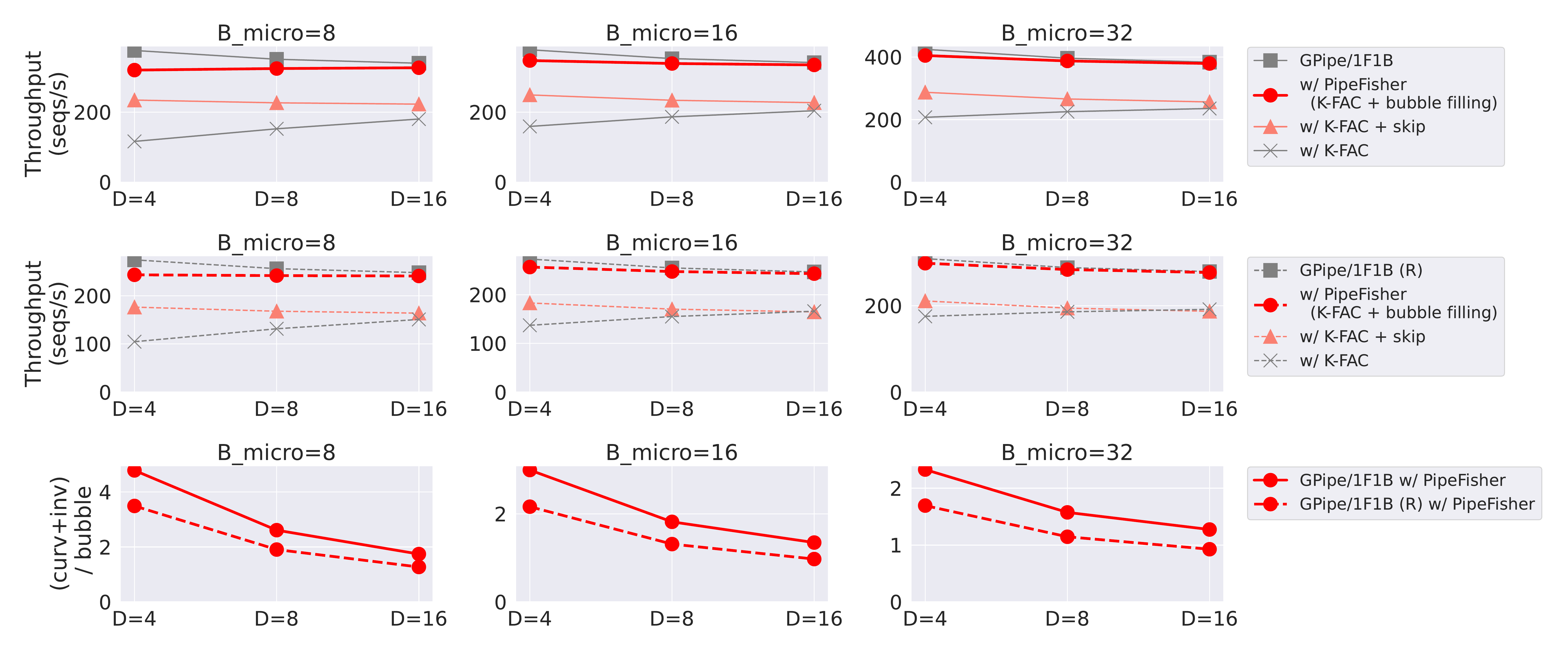}
        \caption{\change{Modeled throughput (sequences/s) and ({\curvature}+{\inversion})-bubble ratio of GPipe and 1F1B (w/ pipeline flush)}}
    \end{subfigure}
    \begin{subfigure}{.8\linewidth}
        \includegraphics[width=\textwidth]{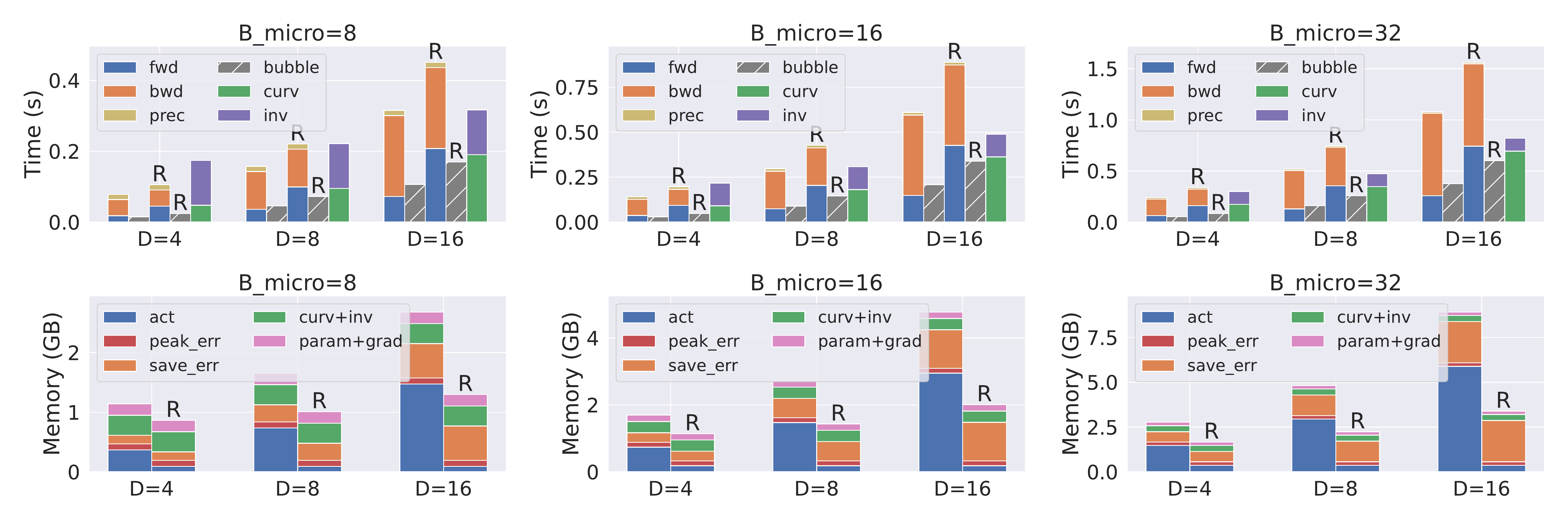}
        \caption{Modeled time per step and memory consumption of Chimera w/ $2$ pipelines}
    \end{subfigure}
    \begin{subfigure}{.8\linewidth}
        \includegraphics[width=\textwidth]{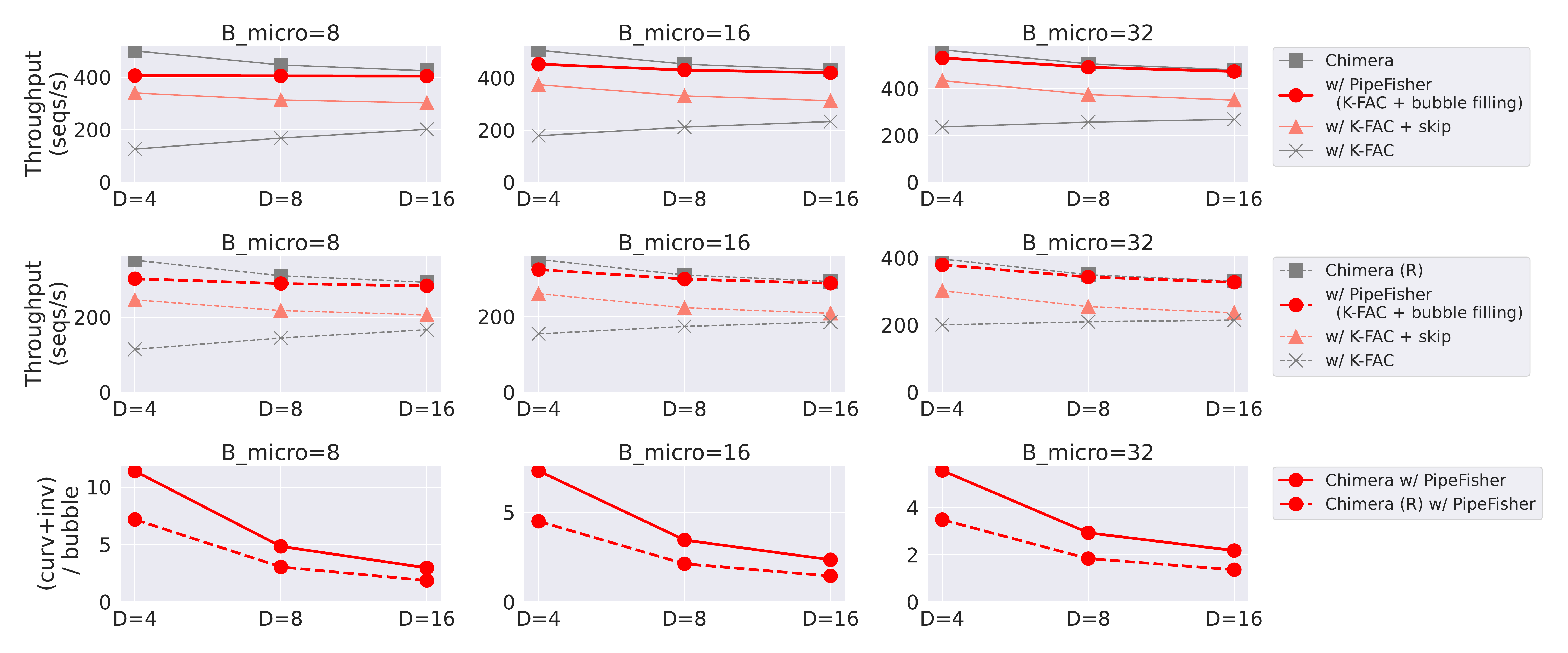}
        \caption{\change{Modeled throughput (sequences/s) and ({\curvature}+{\inversion})-bubble ratio of Chimera w/ $2$ pipelines}}
    \end{subfigure}
    \caption{Performance model for $\pldepth$ BERT-Large blocks (one block per pipeline stage) with $\nmicrobatches=\pldepth$ on an NVIDIA P100. R indicates activation recomputation.}
    \label{fig:bert_large_perf_model}
\end{figure*}

\begin{figure*}
    \centering
    \includegraphics[width=\textwidth]{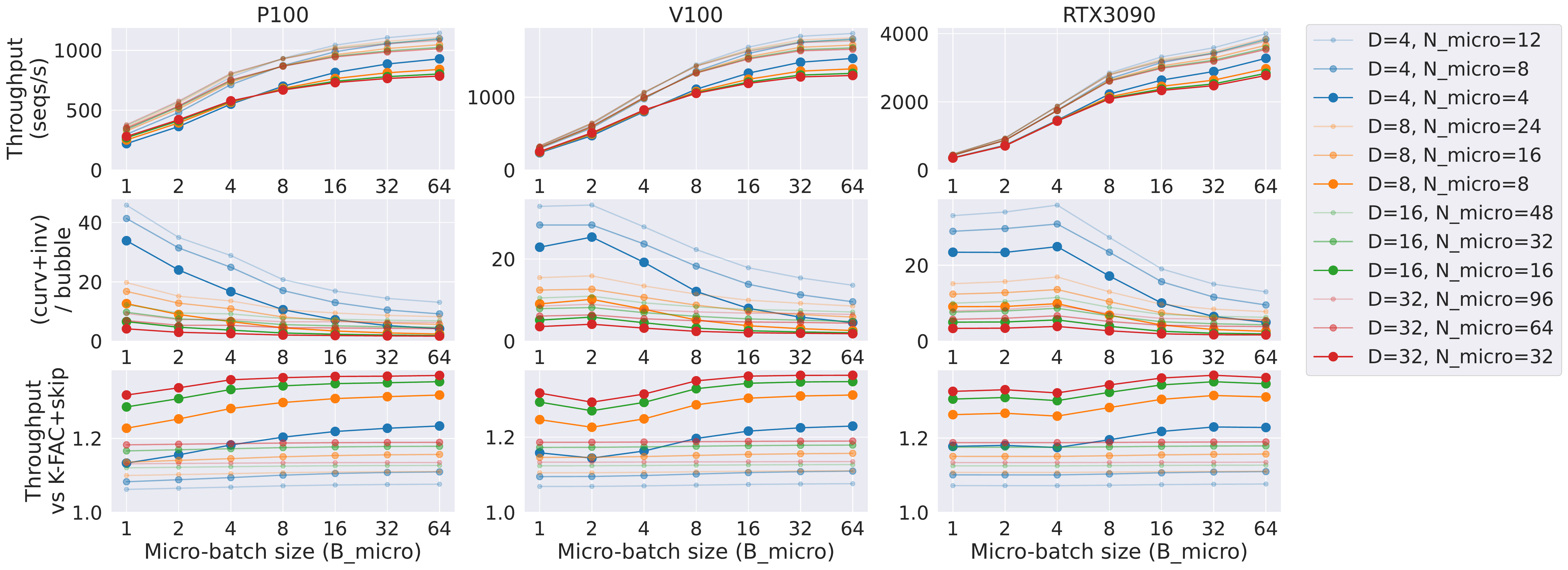}
    \caption{\change{Modeled throughput (sequences/s) and ({\curvature}+{\inversion})-bubble ratio of Chimera w/ $2$ pipelines w/ PipeFisher for $\pldepth$ \textbf{BERT-Base} blocks ($\pldepth\in\{4,8,16,32\}$, one block per pipeline stage) on an NVIDIA P100, V100, and RTX3090.}}
    \label{fig:bert_base_perf_model2}
\end{figure*}

\begin{figure*}
    \centering
    \includegraphics[width=\textwidth]{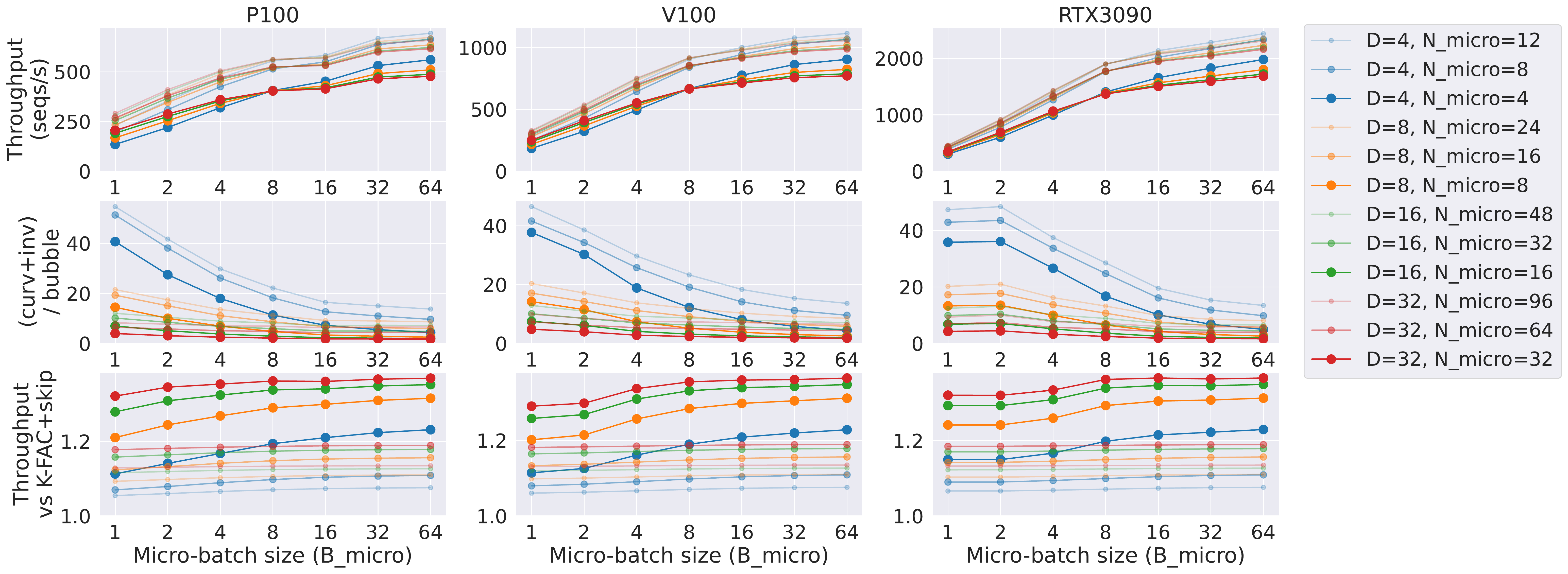}
    \caption{\change{Modeled throughput (sequences/s) and ({\curvature}+{\inversion})-bubble ratio of Chimera w/ $2$ pipelines w/ PipeFisher for $\pldepth$ \textbf{BERT-Large} blocks ($\pldepth\in\{4,8,16,32\}$, one block per pipeline stage) on an NVIDIA P100, V100, and RTX3090.}}
    \label{fig:bert_large_perf_model2}
\end{figure*}

\begin{figure*}
    \centering
    \includegraphics[width=\textwidth]{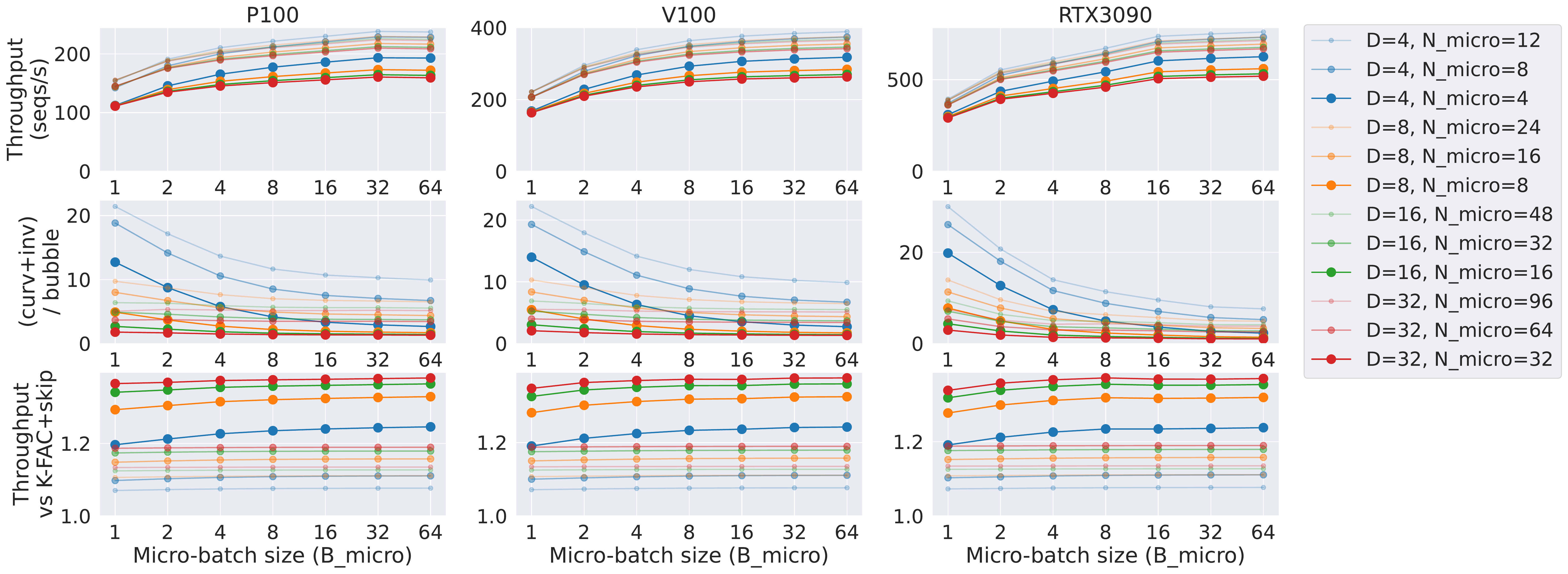}
    \caption{\change{Modeled throughput (sequences/s) and ({\curvature}+{\inversion})-bubble ratio of Chimera w/ $2$ pipelines w/ PipeFisher for $\pldepth$ \textbf{T5-Base} blocks ($\pldepth\in\{4,8,16,32\}$, one block per pipeline stage) on an NVIDIA P100, V100, and RTX3090.}}
    \label{fig:t5_base_perf_model2}
\end{figure*}

\begin{figure*}
    \centering
    \includegraphics[width=\textwidth]{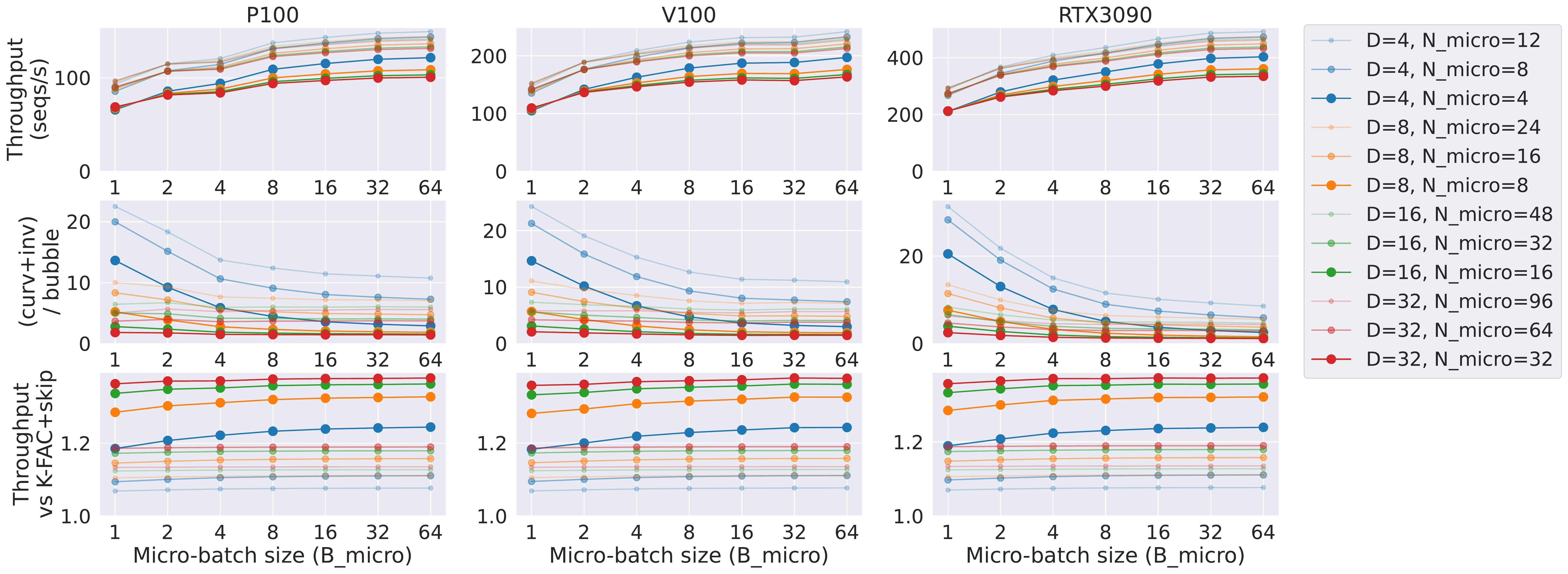}
    \caption{\change{Modeled throughput (sequences/s) and ({\curvature}+{\inversion})-bubble ratio of Chimera w/ $2$ pipelines w/ PipeFisher for $\pldepth$ \textbf{T5-Large} blocks ($\pldepth\in\{4,8,16,32\}$, one block per pipeline stage) on an NVIDIA P100, V100, and RTX3090.}}
    \label{fig:t5_large_perf_model2}
\end{figure*}

\begin{figure*}
    \centering
    \includegraphics[width=\textwidth]{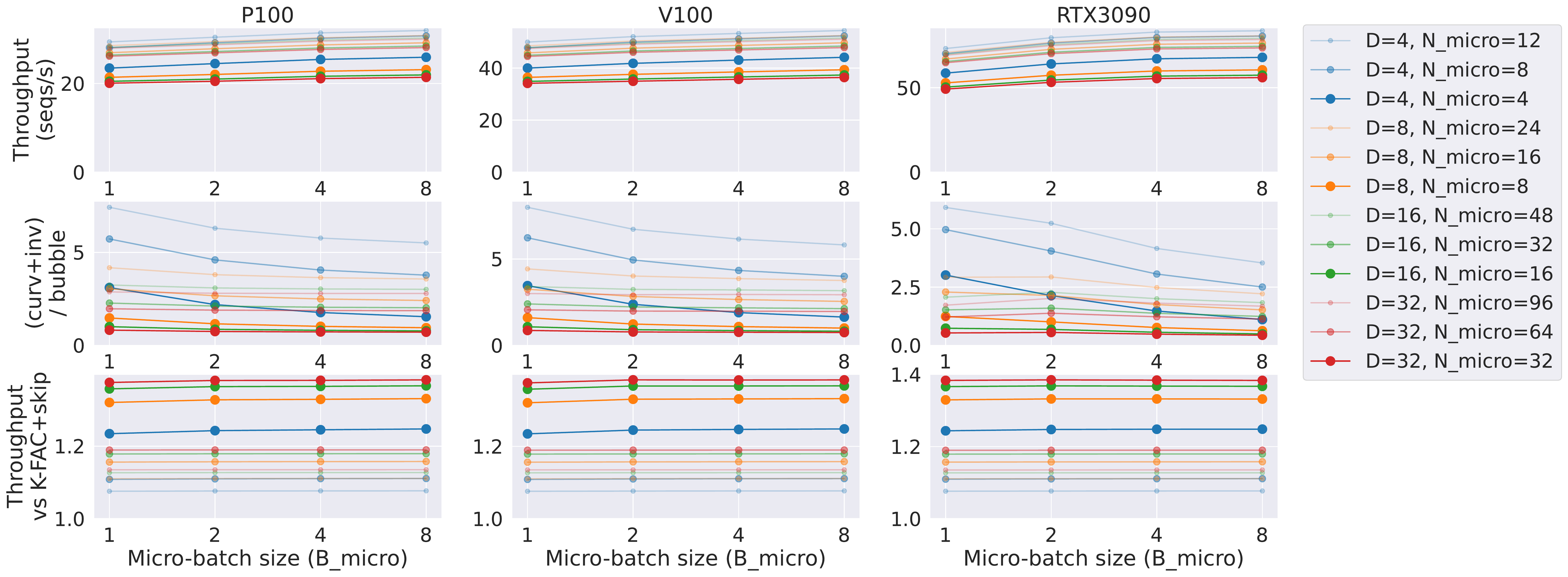}
    \caption{\change{Modeled throughput (sequences/s) and ({\curvature}+{\inversion})-bubble ratio of Chimera w/ $2$ pipelines w/ PipeFisher for $\pldepth$ \textbf{OPT-125M} blocks ($\pldepth\in\{4,8,16,32\}$, one block per pipeline stage) on an NVIDIA P100, V100, and RTX3090.}}
    \label{fig:opt_125m_perf_model2}
\end{figure*}

\begin{figure*}
    \centering
    \includegraphics[width=\textwidth]{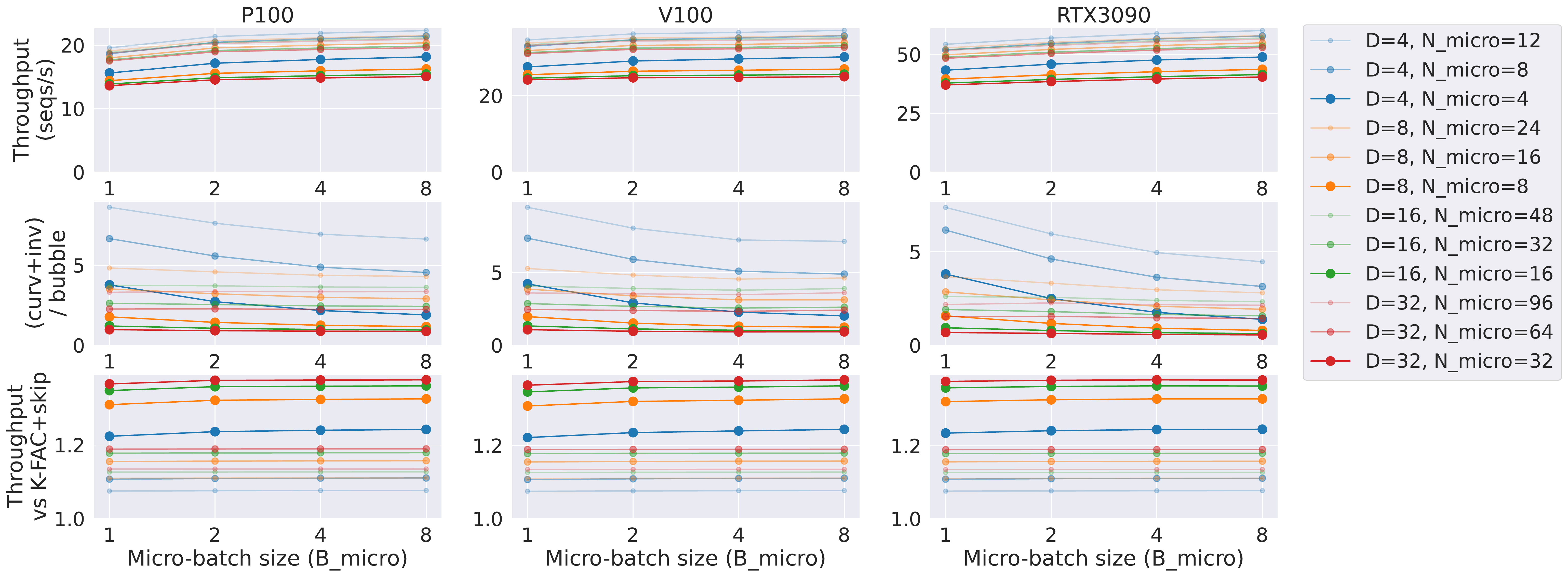}
    \caption{\change{Modeled throughput (sequences/s) and ({\curvature}+{\inversion})-bubble ratio of Chimera w/ $2$ pipelines w/ PipeFisher for $\pldepth$ \textbf{OPT-350M} blocks ($\pldepth\in\{4,8,16,32\}$, one block per pipeline stage) on an NVIDIA P100, V100, and RTX3090.}}
    \label{fig:opt_350m_perf_model2}
\end{figure*}

\end{document}